\documentclass{llncs}
\pdfoutput=1

\usepackage{graphicx,graphics}
\usepackage{color}
\usepackage{url}
\usepackage{listings}
\usepackage{amsmath}
\usepackage{amssymb}
\usepackage{pgfplotstable,booktabs}

\lstdefinelanguage{Comet}
	{ morekeywords={range, enum, int, inc, var, forall, in, while,
	do, if, else, return, void, set, float, Float, move, sum,
	prod, class, tuple, interface, function, operator, implements,
	extends, import, include, native, this, super, union, cross,
	and, or, union, inter, string, Closure, Counter, closure,
	pclosure, Process, break, continue, select, selectMin,
	selectMinAsp, selectMax, selectMaxAsp, selectFirst,
	selectRandom, selectCircular, selectPr, maximizing,
	minimizing, minimize, maximize, minmax, max, argMax, min,
	argMin, new, expr, setof, collect, filter, mapof, all, abs,
	floor, ceil, sqrt, ln, lookahead, IntToString, card, when,
	whenever, foreveron, sleepUntil, neighbor, Boolean, Integer,
	abstract, continuation,
	Continuation,Solution,CPNode,Checkpoint,model, soft, hard,
	call, instanceof, try, tryall, parall, pardo, until, parever,
	onFailure, exploreall, explore, catch, throw, cast, ref, sync,
	shared, dict, implements, switch, case, default, null, Object,
	boolean, bool, true, false, Event, KeyEvent, Condition,
	notify, clearEvents, with, interval, assert, member, thread,
	for, synchronized, by, final, noevent, alias, trail, subject,
	to, using, sin, cos, atan, ln, exp,onDomains,onValues},
	  sensitive=true,
	  basicstyle=\scriptsize,
	  tabsize=2,
	  frame=lines,
	  numbers=left,
	  stepnumber=1,
	  numbersep=1mm,	
	  xleftmargin=6pt,
	  morecomment=[l]{//},
	  morecomment=[s]{/*}{*/},
	  morestring=[b]",
	  morestring=[d]',
	  showstringspaces=false
	}

\lstnewenvironment{Comet}{\lstset{language=Comet}}{}

\newcommand{\comet}{\textsc{Comet}}
\newcommand{\ocp}{\textsc{Objective-CP}}

\newcommand{\remove}[1]{}

\newcommand{\elsdel}[1]{{\color{red}}}

\begin{document}

\title{A Microkernel Architecture \\
for Constraint Programming}

\titlerunning{A Microkernel Architecture for Constraint Programming}

\author{
 L. Michel \and
 P. Van Hentenryck
}

\institute{
 L. Michel 
 University of Connecticut, Storrs, CT 06269-2155\\
 \email{ldm@engr.uconn.edu}
 \and
 P. Van Hentenryck 
 National ICT Australia (NICTA) and the Australian National University \\
 \email{pvh@nicta.com.au}
}
\maketitle

\begin{abstract}
This paper presents a microkernel architecture for constraint
programming organized around a number of small number of core
functionalities and minimal interfaces. The architecture contrasts
with the monolithic nature of many implementations. Experimental
results indicate that the software engineering benefits are not
incompatible with runtime efficiency.
\end{abstract}

\section{Introduction}
Over the years, the constraint programming community has built and
experimented with a variety of system design and implementation
Historically, solvers were embedded in logic programming languages,
\textsc{CHIP} \cite{CHIP}, \textsc{CLP(R)} \cite{CLPR}, and
\textsc{Prolog III} \cite{Colmerauer90} being prime examples. Later
solvers were packaged as libraries hosted in traditional
object-oriented and imperative languages. Examples abound starting
with \texttt{Ilog Solver}~\cite{ILOG}, \texttt{Gecode}~\cite{Gecode09}
and Minion~\cite{Gent06minion:a} for \texttt{C++} and
\texttt{Choco}~\cite{Choco08} or \texttt{JacOP}~\cite{Jacop12} for
\texttt{Java} (to name just a few).  The success of algebraic modeling
languages in the mathematical programming community as well as the
desire to not be constrained by the host language prompted advances in
domain-specific languages, a trend examplified by
\textsc{Oz/Mozart}~\cite{Smolka95}, \textsc{Opl}~\cite{OPL},
\textsc{Salsa}~\cite{SALSA}, and \textsc{Comet}~\cite{CBLS} for
instance. In all cases, the traditional constraint-programming
capabilities are exposed by an API capturing the mantra
\[
CP = Model + Search
\]
\noindent
However, in these implementations, the solver is delivered as a
monolithic piece of software incorporating key notions such as
variables, constraints, events, propagation protocols, propagators, and
search support.  Perhaps even more strikingly, propagation engines
include provisions to deal with decisions variables of many different
types (e.g., \texttt{int,float,set}), possibly replicating the APIs
for each variant.
Orthogonal features (e.g., the support for AC-5, views, and advisors)
further contribute to complexity as APIs must be duplicated to
transmit additional information to the propagators when events of
specific classes occur.
While conceptually simple, this approach does not scale very well. 
Consider the \textsc{Comet} API shown in Figure~\ref{cometapi}. It
blends support for AC-3, AC-5, 
and the ability to convey the index of a variable to a propagator along  
with two types of decision variables. 
The protocol \texttt{AC5Constraint<CP>} in lines 1--13 only captures the
methods that a propagator supporting AC-5 could implement. Line 9, for
instance, specifies method \texttt{valRemoveIdx} which is called whenever
variable $v$ appearing at index $vIdx$ in some array has lost value $val$. 
Similarly, the integer variable \texttt{VarInt<CP>} offers one registration
method for each class of events that the variable is susceptible to raise.
The net result is a large APIs encompassing the union of core capabilities 
for all variable types and event classes.  

\begin{figure}[tb]
\begin{Comet}
interface AC5Constraint<CP> extends Constraint<CP> {
  Outcome<CP> post(Consistency<CP> cl);
  Outcome<CP> valRemove(var<CP>{int} v,int val);
  Outcome<CP> valBind(var<CP>{int} v,int val);
  Outcome<CP> updateMin(var<CP>{int} v,int val);
  Outcome<CP> updateFloatMin(var<CP>{float} v,float val);
  Outcome<CP> updateBounds(var<CP>{int} v);
  ...
  Outcome<CP> valRemoveIdx(var<CP>{int} v,int vIdx,int val);
  Outcome<CP> valBindIdx(var<CP>{int} v,int vIdx,int val);
  Outcome<CP> updateMinIdx(var<CP>{int} v,int vIdx,int val);
  Outcome<CP> updateMaxIdx(var<CP>{int} v,int vIdx,int val);
  ...
} 

native class VarInt<CP> implements Var<CP> {
  ...
  void addMin(Constraint<CP> c);
  void addMax(Constraint<CP> c);
  void addBind(Constraint<CP> c);
  void addAC5(AC5Constraint<CP> c);
  void addAC5Index(AC5Constraint<CP> c,int idx);
  void addAC5Bind(AC5Constraint<CP> c);
  void addAC5BindIndex(AC5Constraint<CP> c,int idx);
  void addAC5MinIndex(AC5Constraint<CP> c,int idx);
  ...
}
\end{Comet}
\caption{\textsc{Comet} APIs for AC-5 constraints and integer
 variables.}
\label{cometapi}
\end{figure}

This paper examines a possible alternative to such a design: It
describes the microkernel of the \ocp{}~\cite{ObjectiveCP} system, a
new constraint programming system that refines the core mantra to
\[
Optimization\:\: Program = Model + Search + Solver
\] 
\ocp{} isolates all the responsibilities associated with model
manipulation, rewriting, specialization and reformulation in the
modeling component. This factorizes key capabilities that are reusable
by multiple solvers and promotes the idea of model combinators
described in~\cite{ModelCombinators13}.
\ocp{} also supports search facilities through a technology-neutral
search library featuring combinators and a seamless symbiosis with the
host language, i.e., \textsc{Objective-C}. The search
library~\cite{Search13} makes it possible to specify and execute
complex search procedures with minimal effort and delivers
competitive performance.
The \ocp{} constraint-programming solver embraces the idea of a
microkernel architecture inspired by recent develoments in operating
systems.  The constraint engine features a number of small components
with parsimonious interfaces. Microkernel architectures have become
very popular in operating systems as they favor extensibility and
maintenance, and make proofs of correctness easier.
The constraint-programming engine underlying \ocp{} delivers a truly
modular architecture where variables, domains, and constraints all
remain completely external to the microkernel.  Each computational
domain (e.g., booleans, reals, integers, sets) becomes a service on
top of the microkernel, providing the necessary infrastructure for
propagation-driven inferencing. In particular the microkernel isolates
the propagation logic and events protocols from the variables and
constraint definitions. This architecture is therefore a departure
from the monolithic organization prevalent in modern solvers. It is
designed to encourage the construction of separate libraries each
featuring different domains, variable representation (e.g.,
finite-domains, intervals, sets, MDDs) and constraints.

The rest of the paper is organized as
follows. Section~\ref{sec:overview} provides a brief overview of
\ocp{} and the vision for the system.  Section~\ref{sec:ukernel}
covers the functionalities of the microkernel.
Section~\ref{sec:fdservice} discusses how to use the microkernel to
implement a finite-domain solver. Section~\ref{sec:empirical} offers
empirical evidence of the platform capabilities and
Section~\ref{sec:ccl} concludes the paper.

\section{Overview of \ocp{}}
\label{sec:overview}

\begin{figure}[t]
\begin{center}
\includegraphics[width=4.5in]{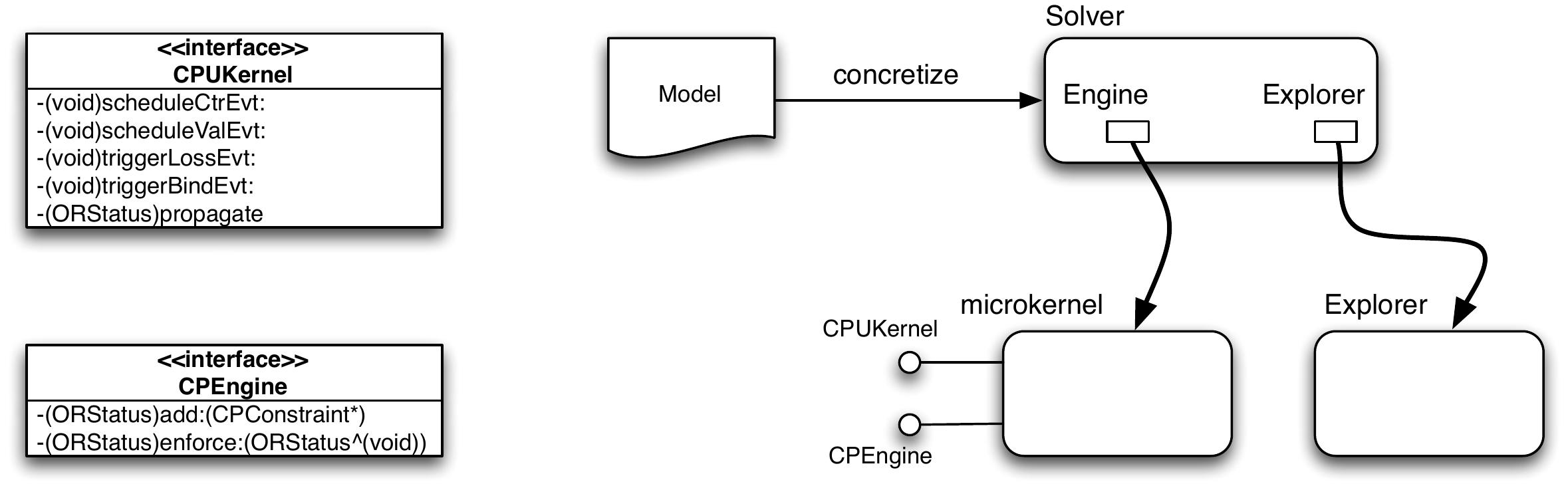}
\end{center}
\vspace{-4mm}
\caption{The Structure of a CP Program in \ocp{}.}
\label{fig:cpprogram}
\end{figure}

The design of \ocp{} takes the view that
\begin{quote}
Optimization Program = Model + Search + Solver
\end{quote}
or, in other words, that an optimization program consits of a model, a
search, and an underlying solver. The overall architecture is
illustrated in Figure~\ref{fig:cpprogram} where a model is concretized
into a solver.  The solver is the composition of an engine
(responsible for the representation of constraint, variables, and the
inferencing) and an explorer (responsible for technology-neutral search
capabilities). The kernel itself implements the two interfaces shown on
the left-hand side to drive the propagation (the \texttt{CPUKernel})
and to register constraints and variables (the \texttt{Engine}).

\subsection{The Vision underlying \ocp{}}

\paragraph{Models} 
Models are first-class objects in \ocp{}. They follow the style of
constraint programming and are solver-independent. This allows for
easy experimentation with different technologies and smooth
hybridizations \cite{ModelCombinators13,Fontaine12}. Models can be
transformed and refined through a sequence of operators that, for
instance, replace algebraic equations with lower-level elementary
constraints. The final model is then concretized into a specific
solver to obtain an optimization program, (e.g., a constraint program
or a mixed-integer program). Given a model $M$, a sequence of model
operators $\tau_0,\cdots\tau_k$, and a concretization function
$\gamma_T$ for a solver technology $T$, the application
\[
\gamma_T(\tau_k(\cdots \tau_1(\tau_0(M))\cdots))
\]
derives a concrete program based on technology $T$. The resulting
program can be solved using a black-box search or a dedicated search
procedure expressed in terms of the model variables present in $M$.

\paragraph{Search} 
Search procedures in \ocp{} are specified in terms of high-level
nondeterministic constructs, search combinators, and node selection
strategies, merging the benefits of search controllers and
continuations \cite{CPAIOR05} on the one hand and compositional
combinators (e.g., \cite{Stuckey13}) on the other hand. The search
language is generic and independent of the underlying
solver. Naturally, search procedures call the underlying engine for
adding constraints, binding variables, and querying the search state.

\paragraph{Engine} The underlying engines can leverage any combinatorial
optimization technology ranging from linear programming and integer
programming to constraint programming  and constraint-based local search. The
role of the engine is to isolate the  state representation and inferencing
capabilities. This paper focuses on a microkernel architecture for the
 inference engine focused on ``traditional'' (finite-domain) constraint
programming. 

\subsection{First-Order Functions}
\ocp{} leans heavily on its host language (Objective-C) to implement
its capabilities.  In particular, it relies on closures and
first-order functions.  In Objective-C, it is straightforward to turn
any C block into a first-order function. For instance, the fragment
\begin{lstlisting}
[S enumerateWithBlock: ^(id obj) {
	printf("object: %@\n",[obj description]);
}];
\end{lstlisting}
uses an Objective-C block to produce a first-order function capable of
printing an object ($obj$). The first-order function is then used to
visit a set $S$ and print its entire content. The syntax of
Objective-C block is reminiscent of the syntax for function pointers
in $C$: The caret symbol indicates a block creation.

\subsection{An \ocp{} Primer}
\begin{figure}[t]
\begin{lstlisting}
id<ORModel> m = [ORFactory createModel];

// data declarations and reading

id slab = [ORFactory intVarArray: m range: SetOrders domain: Slabs];
id load = [ORFactory intVarArray: m range: Slabs domain: Capacities];

[m add: [ORFactory packing: slab itemSize: weight load: load]];
for(ORInt s = Slabs.low; s <= Slabs.up; s++)
   [m add:[Sum(m,c,Colors,Or(m,o,coloredOrder[c],[slab[o] eq:@(s)])) leq:@2]];
[m minimize: Sum(m,s,Slabs,[loss elt: load[s]])];
 
id<CPProgram> cp = [ORFactory createCPProgram: m];

[cp solve: ^{
   for(ORint i = SetOrders.low; i <= SetOrders.up; i++) {
      ORInt ms = max(0,[cp maxBound: slab]);
      [cp   tryall: Slabs suchThat: ^bool(ORInt s) { 
      	               return s <= ms+1 && [cp member:s in:slab[i]];
      	            } 
                in: ^void(ORInt s) { [cp label: slab[i] with: s];}
         onFailure: ^void(ORInt s) { [cp  diff: slab[i] with: s];}
      ];
   }
}];
\end{lstlisting}
\vspace{-4mm}
\caption{The Steel Mill Slab Problem in \ocp{}.}
\label{fig:slab}
\end{figure}

This section is a brief primer to \ocp{}.  Consider the Steel Mill
Slab problem as an example. An abridged\footnote{The data reading part
  of the program is omitted for brevity.} \ocp{} implementation is
shown in Figure~\ref{fig:slab}.
In this program, line 1 first creates a model $m$, while lines 5 and 6
create two decision variable arrays $slab$ and $load$ to represent the
slab assignment and the load of each slab. Lines 8--11 state the model
constraints, i.e., a global packing constraint and a coloring
constraint for each slab, and the objective function (line 11) which
minimizes the total loss.
The model is not specific to constraint programming and Line 13
transforms and concretizes this high-level model into a constraint
program. While $m$ is clearly constructed with algebraic expressions,
the microkernel underlying the CP engine operates on a rewriting of
the model featuring exclusively low-level and global constraints. To
be more precise, line 13 is equivalent to

\lstset{language=[Objective]C,mathescape=true,numbers=none,numbersep=1mm,xleftmargin=2pt,frame=lines,
tabsize=2,
basicstyle=\scriptsize\ttfamily}
\begin{lstlisting}
cp = $\gamma_{CP}(\tau_0$(m)$)$;
\end{lstlisting}
\lstset{language=[Objective]C,mathescape=true,numbers=left,numbersep=1mm,xleftmargin=6pt,frame=lines,
tabsize=2,
basicstyle=\scriptsize\ttfamily}

\noindent where $\tau_0(m)$ is a flattening operator that creates a
new model based on $m$ in which all relations and expressions have
been replaced by basic constraints. $\gamma_{CP}(M)$ is a
concretization function that associates a concrete variable with each
modeling variable in $m$, as well as a propagator with every modeling
constraint in $m$. The concretized model is then \textit{loaded} into
the engine of a constraint-programming solver.
Finally, lines 15--25 implement the search and rely on the
\texttt{tryall:} combinator to specify the non-deterministic
choices. It is worth noticing how \ocp{} blends the control primitive
of the host language with search combinators in a completely
transparent and fully compositional way.
From an \ocp{} end-user standpoint, the entire search is expressed in
terms of the decision variables from $m$. Method calls such as
\texttt{label:with:} and \texttt{diff:with:} are exclusively
manipulating model variables, yet they are automatically recasted in
term of concrete variables and delegated to the underlying engine.
Indeed, the call in the closure on line 21 becomes
\begin{lstlisting}
[[cp engine] add: $\gamma_{CP}(\tau_0(\mbox{\tt slab[i]} == \mbox{\tt s} ))$];
\end{lstlisting}
where the abstract constraint \texttt{slab[i] == s} is flattened with
$\tau_0$ and concretized with $\gamma_{CP}$ before being transmitted
to the underlying engine.

\subsection{A Solver Interface}

A solver is the composition of an \textit{explorer} responsible for
the search and an \textit{engine} responsible for inferencing.  The
engine implements two distinct interfaces. The {\em engine} interface
is the solver-facing part of \ocp{} needed for concretizing models: It
offers the necessary capabilities to register variables, constraints,
and objective functions.  The {\em microkernel} interface is used by
system developers when building new propagators: It offers the
functionalities for propagation-related activities such as the
dispatching of events.

\begin{figure}[tb]
\vspace{-4mm}
\noindent
\begin{minipage}{1\textwidth}
\noindent
\begin{lstlisting}
@protocol CPEngine
-(ORStatus)      add: (id<CPConstraint>) c;
-(void) setObjective: (id<ORObjective>) obj;
-(ORStatus)  enforce: (ORStatus^(void)) cl;
@end
\end{lstlisting}
\end{minipage}
\vspace{-4mm}
\caption{A constraint programming \texttt{CPEngine} Interface.}
\label{fig:engine}
\vspace{2mm}
\end{figure}

\paragraph{The Engine}

The interface shown in Figure~\ref{fig:engine} is the visible tip of
the iceberg for \textit{microkernel users.} It is the API
needed to load constraints and objective functions.  It offers only
three methods to register a concrete constraint over concrete
variables (line 2); to set the objective function via the
\texttt{setObjective:} method (line 3); and to execute an arbitrary
closure and propagate its effects with the \texttt{enforce:} method
(line 4).
The closure $cl$ passed to \texttt{enforce:} returns a status (i.e.,
an element in the set $\{fail,suspend,succeed\}$) to report the
propagation outcome.

\paragraph{Constraints}

The \texttt{CPConstraint} protocol used by the engine is shown below
\begin{lstlisting}
@protocol CPConstraint
-(ORUInt) getId;
-(void)    post;
@end
\end{lstlisting}
It only requires that a constraint carries a unique identifier and
responds to a \texttt{post} request.  Naturally, constraints have
additional methods but these are not mandated by the microkernel.

\paragraph{Objective}

The \texttt{ORObjective} protocol is shown below alongside a protocol
to describe the value of an objective function.

\begin{lstlisting}
@protocol ORObjective
-(id<ORObjectiveValue>) value;
@end

@protocol ORObjectiveValue 
-(id<ORObjectiveValue>) best: (id<ORObjectiveValue>) other;
-(ORInt)             compare: (id<ORObjectiveValue>) other;
@end
\end{lstlisting}

At its most abstract level, one can query an objective to retrieve its
current value represented as an abstract object as well.  Note how
\texttt{ORObjectiveValue} instances can be compared to pick the best
value. Naturally, one should only compare objective values issued by
the same objective function, so that the optimization direction is
correctly taken into account. For instance, a call to \texttt{best:}
on an objective value $a$ receiving an objective value $b$ returns the
best overall objective value. If the objective function was a
minimization over the real, the returned result encapsulates a value
ranging over the reals and retains the knowledge that it is coming
from a minimization.

\section{The Microkernel Architecture}
\label{sec:ukernel}

The purpose of the microkernel is to act as a relay for messages
pertaining to the propagation of constraints. The key challenge is to
design a small set of capabilities to suppot variables, constraints,
and propagation techniques of different forms. While it is always
possible to define a kernel offering the union of all the required
capabilities, the approach does not scale and is truly intrusive when
designing new classes of constraints with different messaging
requirements as illustrated in the introductory example with the
\textsc{Comet} API in Figure~\ref{cometapi}.

\begin{figure}[t]
\vspace{-4mm}
\noindent\begin{minipage}{1\textwidth}
\begin{lstlisting}
@protocol CPUKernel
-(void) scheduleClosureEvt: (id<CPClosureEvent>) list;
-(void) scheduleValueEvt: (id<CPValueEvent>) list  with:(id)e;
-(void) triggerLossEvt: (id<CPTriggerMap>)tmap with:(id)e;
-(void) triggerBindEvt: (id<CPTriggerMap>)tmap;
-(ORStatus) propagate;
@end 
\end{lstlisting}
\end{minipage}
\vspace{-4mm}
\caption{A constraint programming $\mu$-Kernel Interface.}
\label{fig:ukernel}
\end{figure}

The constraint-programming microkernel of \ocp{} presents an
alternative design based on a minimalistic API with two capabilities:
\begin{itemize}
\item scheduling events
\item propagating events.
\end{itemize}

\noindent The API of the microkernel per se is shown in
Figure~\ref{fig:ukernel}. Line 2 descibes the method for scheduling
constraint events, Line 3 the method for scheduling \textit{value}
events, Line 4 the method for scheduling all triggers associated with
a value loss, and line 5 the method for scheduling all triggers
associated with a value binding.  Finally, line 6 is the method needed
to trigger the inferencing.
These APIs only require a handful of other protocols that embody the
concept of \textit{event lists and maps}. Namely, they reference
\texttt{CPClosureEvent}, \texttt{CPValueEvent}, and \texttt{CPTriggerMap}.
In the following section, responding to \textit{events} is best
understood as executing an arbitrary piece of code represented by a
function.  The exact nature of the response is discussed in
Section~\ref{sec:evtlist}.

\subsection{Propagation Preliminaries}
\label{sec:prelim}

\paragraph{Events}

The concept of \textit{event} is the cornerstone of the
microkernel. Events are the vehicle of choice to relay information and
takes a very abstract form that is independent of the nature of the
variables involved. In this paper, \textit{events} are closures in a
functional-programming sense. Namely, they are blocks of code that
capture the computational state at the time of their definition and
are wrapped in a first-order function (of type $\mbox{\tt void}
\rightarrow \mbox{\tt void}$) that can be saved, called, or passed to
other functions.

\paragraph{Priority Space}

The microkernel supports event priorities, using a range of numerical
values $0..P$, where $P$ is the highest priority and $0$ is the
lowest. Two priorities in this range have special statuses. Priority
$0$ is \textit{always} dispatched regardless of the outcome of the
propagation, i.e., even in case of a failure.  Priority $P$ is
reserved for value-driven events. The remaining priorities ($1..P-1$)
are available for general use. The existence of a special priority $0$
may sound surprising at first.  However, it is the ideal vehicle to
implement key functionalities in a non-intrusive way. To illustrate,
simply consider black-box search heuristics such as \textsc{Ibs} and
\textsc{Abs}. Both necessitate that, at the end of a propagation cycle
and irrespective of the outcome, variable statistics be updated for
every variable involved in the fixpoint (e.g., for \textsc{Abs}
search, one must update the activity of the variables that
participated in the fixpoint computation). In a traditional kernel,
such a support requires the instrumentation of the solver to invoke,
at the end of the fixpoint, the code fragment responsible for updating
those statistics.  An \textit{always priority} (i.e., priority 0)
solves this problem. Indeed, one can simply attach a daemon with every
variable and schedule it at priority 0. When the daemon runs at the
end of the propagation, it updates the statistics stored in the
implementation of \textsc{Ibs} or \textsc{Abs}.  Priority 0 can also
be useful for implementing visualizations where some redraw must be
done regardless of the propagation outcome and are driven by the
variables touched during the propagation.  

\paragraph{Queues}

The microkernel of \ocp{} is responsible for dispatching
\textit{events} arising as a result of the propagation of constraints.
To this end, it relies on an array of $P$ queues in which $Q_i$ refers
to a queue at priority $i$. 

\subsection{The Propagation Engine}

While an OS microkernel is tasked with \textit{continuously}
dispatching messages to the processes it manages, the \ocp{}
microkernel only dispatches accumulated messages at specific points
during the execution. This section is concerned with the dispatching
process alone and it presents the implementation of the
\texttt{propagate} method of Figure~\ref{fig:ukernel}.

\begin{figure}[t]
\begin{lstlisting}
-(ORStatus)propagate {
	BOOL done = NO;
	return tryfail(^{
		while (!done) {
			p = $\max_{i=1}^{P} i \cdot (Q_i \neq \emptyset)$
			while ($p \neq 0$) {
				call(deQueue($Q_p$));
				p = $\max_{i=1}^{P} i \cdot (Q_i \neq \emptyset)$;
			}
			done = $Q_p = \emptyset$;
		}
		while ($Q_0 \neq \emptyset$) 
			call(deQueue($Q_0$));
		return ORSuspend;
	}, ^ {
		while ($Q_0 \neq \emptyset$) 
			call(deQueue($Q_0$));
		return ORFail;
	});
}
\end{lstlisting}
\vspace{-4mm}
\caption{The {\sc Propagate} Method in \ocp{}.}
\label{dispatch:ac3}
\end{figure}

\paragraph{The Propagation Loop}

The dispatching algorithm is shown in Figure~\ref{dispatch:ac3}: It
processes each non-empty queue in turn from the highest ($P$) to the
lowest ($1$) priority. Line 5 finds the index of the highest priority
queue with some events. Lines 6--9 pick the first highest priority
event, dispatch it (line 7) and carry on until $p=0$ which indicates
that all queues in the $1..P$ range are empty.  Finally, lines 12--13
unconditionally execute all the events held in $Q_0$. As is customary,
the dispatching of messages may schedule additional events that will
be handled during this cycle. Since individual events are represented
by closures of the form $B : \mbox{\tt void} \rightarrow \mbox{\tt
  void}$, dispatching an event is modeled by a simple instruction
$call(B)$ that executes closure $B$.
%

\paragraph{Handling Inconsistencies}

Producing an elegant propagator implementation can be a challenge with
modern constraint-programming solvers. Variable updates triggered by a
propagator can lead to domain wipe-outs, revealing an inconsistent
computation state (aka a failure). Programmers are therefore expected
to lace the propagator implementation with failure checks and to abort
the propagation when a failure is encountered. Each propagator must
also return a suitable status, indicating whether the propagation
failed.

\ocp{} relies on an alternative design and relies on {\it native
  exceptions} to report failures. The block spanning lines 4--14 in
Figure~\ref{dispatch:ac3} invokes closures that capture the logic of
propagators and can potentially induce failures. It is therefore
captured in a closure and passed alongside a second closure (lines
16--18) to the utility function {\tt tryfail}. The semantics of {\tt
  tryfail($b_0$,$b_1$)} is similar to a try-catch block, i.e., it can
be understood as the rewriting: \\

\begin{lstlisting}
try {
	$b_0$
} catch (FailException* fx) {
	$b_1$
}
\end{lstlisting}
\noindent that executes $b_0$ and transfers control to $b_1$ in case
a failure exception is raised. 

Exceptions, however, are meant to alter the control flow in {\it rare
  and exceptional} conditions. As a result, the implementation of
exceptions, e.g., the {\tt libunwind} library in {\tt C++}, induces a
negligible overhead when executing {\tt try} blocks but incurs a more
significant cost when throwing and unwinding the stack to catch and
handle the exception. Failures in constraint programming are rather
frequent however and such an implementation would produce
non-negligible slowdowns.

To remedy this potential limitation, \ocp{} implements a low-cost
exception mechanism. Function {\tt tryfail} is not implemented in
terms of native exceptions but relies on continuations to achieve the
control-flow transfer. Its pseudo-code is shown in
Figure~\ref{fun:tryfail}. The implementation is reentrant and
thread-safe. Line 1 declares a thread-local variable pointing to a
resume continuation {\tt failPoint}.  {\tt tryfail} starts by saving,
in local storage, the current resumption point in line 4. Line 5
creates a lightweight continuation representing the catch block.  If
the fresh resume continuation was never called (line 6), this is the
equivalent of the try block and lines 7--10 execute $b_0$ after
installing the catch handler $k$ in thread-local storage {\tt
  failPoint}.  If $b_0$ succeeds, the previous catch handler is
restored in line 9 and executions leaves the {\tt tryfail}. If an
``exception'' is raised (via a call to {\tt fail} shown in lines
16--19), the current continuation in {\tt failPoint} is called and the
control flow reaches line 6 again, but this time the number of calls
is positive and the block in lines 12--13 executes. This final step
also restores the previous resume continuation and proceeds with a
call to $b_1$. Observe that this implementation does not use the full
power of a continuation, merely its ability to alter the control-flow
and shrink the system stack. Therefore, it can be implemented in term
of the classic $C$ functions \texttt{setjmp} and \texttt{longjmp} for
an even lower overhead.

\begin{figure}[t]
\begin{lstlisting}
static __thread Continuation* failPoint = NULL;

ORStatus tryfail(ORStatus (^b0)(),ORStatus (^b1)()) {
	Continuation* ofp = failPoint;
	Continuation* k = [ORContinuation takeContinuation];
	if (k.nbCalls == 0) {
	   failPoint = k;
	   ORStatus rv = b0();
	   failPoint = ofp;
	   return rv;
	} else {
	   failPoint = ofp;
	   return b1();
	}
}
void fail()
{
  [failPoint call];
}
\end{lstlisting}
\vspace{-4mm}
\caption{The \texttt{tryfail} Utility Function.}
\label{fun:tryfail}
\end{figure}


\subsection{Dispatching Events}
\label{sec:evtlist}

This section discusses how to dispatch events for propagation, i.e.,
how the various closures are inserted in the propagation queues and
where they come from. Broadly speaking, the microkernel handles three
classes of events:

\begin{description}
\item[Closure Events:] These events simply insert a closure in a queue when responding to an event;
\item[Value Events:] These events insert a closure obtained from a first-order function and a value;
\item[Trigger Events:] These events associate closures with values and
  can dispatch all the closures associated with a specific value.
\end{description}

\noindent
For finite-domain constraint programming, closure events are typically
used for constraint-based propagation in the style of AC-3. Value
events are typically used for implementing AC-5 style of propagation,
e.g., to propagate the fact that a variable has lost a value. Trigger
events can be used to implement the concept of watched literals and
the ``dynamic and backtrack stable triggers'' described in
\textsc{Minion}~\cite{Gent06}. The microkernel provides abstractions
representing lists or maps of these events. These are used outside the
microkernel for dispatching events as
appropriate. Section~\ref{sec:fdservice} illustrate their use in the
finite-domain service of \ocp{}.

\paragraph{Closure-Event Lists} Closure-Event lists are simply a set of closures
and their associated priorities.

\begin{definition}[Closure-Event List]
  A closure-event list is a list 
\[
l = (\langle f_0,p_0\rangle , \cdots  , \langle f_k,p_k\rangle)
\]
where $f_i :\mbox{\tt void}\rightarrow\mbox{\tt void}$ is a closure
and $p_i \in 0..P-1$ denotes a priority.
\end{definition}
The \texttt{CPClosureEvent} protocol in \ocp{} is used to represent
\textit{closure-event lists} which are ubiquitous in the
implementation: They are used for instance for propagating
constraints, in which case the closure invokes method {\tt propagate}
on the constraint. Closure-event lists are dispatched using method
\begin{lstlisting}
-(void) scheduleClosureEvt: (id<CPClosureEvent>) list;
\end{lstlisting}
whose specification is given by the following definition.

\begin{definition}[Closure-Event Scheduling]
\label{def:cstrevt_sched}
Given a closure-event list \\
$l = (\langle f_0,p_0\rangle, \cdots, \langle
f_{k-1},p_{k-1}\rangle)$, scheduling $l$ amounts to enqueueing each
function in its respective queue, i.e.,
\[
Q_{p_i} = \mbox{\tt enQueue}(Q_{p_i}, f_i)  \:\:\: (0 \leq i \leq k-1).
\]
\end{definition}

\paragraph{Value-Event Lists} Value-Event list contains unary
first-order functions used to respond to generic events that are
instantiated with specific values. A typical example in finite-domain
constraint programming is a propagation event dispatched every time a
value is removed from the domain, in which case the first-order
function expects the removed value as argument.

\begin{definition}[Value-Event List]
  A value-event list is a list $l = ( f_0 , \cdots , f_k )$ where each
  entry $f_i$ is a first-order function of signature ${\cal E} \rightarrow \mbox{\tt void}$.
\end{definition}

\noindent
The exact nature of the event is captured by the opaque datatype
${\cal E}$. Consider, for example, a finite-domain solver over integer
variables where the domain of a variable $x$ is a set
$D(x)=\{0,1,2,3,\cdots,k-1\}$ of $k$ distinct integers. In this case,
${\cal E} = \mathbb{Z}$ and, whenever $v$ disappears from $D(x)$, a
closure $f : \mbox{\tt int} \rightarrow \mbox{\tt void}$ must be
executed on value $v$ to relay the loss to the interested propagator.
The \texttt{CPValueEvent} protocol of \ocp{} is used to represent
\textit{value-event lists}. 

Dispatching a value-event amounts to creating a closure that applies
the first-order function on the arguments. More precisely, value-event
lists are dispatched using the method
\begin{lstlisting}
-(void) scheduleValueEvt: (id<CPValueEvent>) list with:(id)e;
\end{lstlisting}
whose specification is given by the following definition.

\begin{definition}[Value-Event Scheduling]
  Given a value event $e \in {\cal E}$ and a value-event list
  $l=(f_0,\cdots,f_k)$ with $f_i : {\cal E} \rightarrow \mbox{\tt
    void}$, scheduling $e$ amounts to adding the 0-ary closure
  $\lambda.f(e)$ into $Q_P$, i.e.,
\[
Q_P = \mbox{\tt enQueue}(Q_P,\lambda .f(e))   \:\:0 \leq i \leq k-1
\]
\end{definition}
Observe that $\lambda.f(e)$ is a closure whose role is to evaluate
$f(e)$.\footnote{It is given in lambda-calculus
  notation~\cite{Barendregt84} for simplicity.} Clearly, this closure
delays the evaluation of function $f$ on $e$ until the event is pulled
from the queue and propagated.\footnote{An alternative implementation
  can easily store in $Q_P$ objects representing pairs of the form
  $\langle f_i,e\rangle$ and delegate to a method of the pair the task
  to evaluate $f_i(e)$ when the pair is dequeued from $Q_P$. To a
  large extent, this is an implementation detail.}. The above
definition can easily be extended with priorities although, in
general, value-events are not time-consuming and used to perform
simple propagation steps and/or to update some internal data
structures.

\paragraph{Trigger-Events Maps}

Triggers are used in variety of constraint-programming systems (e.g.,
~\cite{CCFD,Gent06}) and can serve as a basis for implementing
generalizations of watched literals in SAT~\cite{Moskewicz:2001}.
\ocp{} supports a general form of trigger-events, making them independent
of finite-domain constraint programming. 

To illustrate the type of propagation supported by triggers, consider
for instance a constraint
\[
\sum_{i=0}^n b_i \geq c
\]
where each $b_i \: i \in 0..n$ is a boolean variable. 
The idea is that a propagation algorithm only needs to listen to $c+1$
variables which can be a substantial saving when $c$ is much smaller
than $n$.  Assume that the propagator is listening to $c+1$ variables
$b_k$ which all satisfy $1 \in D(b_k)$. When such a variable $b_i$
loses value $1$, the propagator searches for a replacement support
$b_j$ among the non-watched variables. If such a support is found, the
propagator starts listening to $b_j$ instead of to $b_i$. If no such
support is found, the $c$ variables that still have $1$ in their
domains must be equal $1$.

\begin{definition}[Trigger Map]
\label{def:triggermap}
A trigger map 
\[
T_m = \{v_i \mapsto \{f_{i,0},\ldots,f_{i,i_k}\} \ \mid
\ 1 \leq i \leq n \}
\]
is a dictionary associating value $v_i$ with closures
$f_{i,0},\ldots,f_{i,i_k}$ $(1 \leq i \leq n)$. We use $T_m(v_i)$ to
denote the set $\{f_{i,0},\ldots,f_{i,i_k}\}$, $dom(T_m)$ the set of
values $\{v_1,\ldots,v_n\}$, and $T_m[w \mapsto S]$ the map $T_m$
where the closures associated with value $w$ are replaced by $S$.
\end{definition}

\begin{figure}[t]
\begin{lstlisting}
@protocol CPTriggerMap <NSObject>
-(id<CPTrigger>) add: (void^()) f forValue:(id) w;
-(void) dispatch: (id) w;
-(void) addTrigger: (id<CPTrigger>) t;
-(void) removeTrigger: (id<CPTrigger>) t;
@end
\end{lstlisting}
\vspace{-4mm}
\caption{Triggers}
\label{fig:trig}
\end{figure}

\noindent The \texttt{CPTriggerMap} protocol in Figure~\ref{fig:trig}
offers four methods to build and use a trigger map.  The
\texttt{add:forValue:} registers a response closure for a
value, i.e., a call \texttt{add:}$f\:\:$\texttt{forValue:}$w$
updates $T_m$ as follows
\[
\left\{\begin{array}{ll}
T_m \ \cup \ \{w \mapsto \{f\}\} &\mbox{ if } w \notin dom(T_m); \\
T_m[w \mapsto \{f\} \cup T_m(w)] &\mbox{ if } w \in dom(T_m).
\end{array}\right. 
\]
It is worth noting that triggers do not refer to variables and provide
a generic capability of the microkernel.  Method \texttt{dispatch:} is
used to execute the triggers associated with a value.

\begin{definition}[Trigger-Event Scheduling]
\label{def:vtsched}
Given a trigger map $T_m$ and a value $v$, scheduling the trigger event
for $t_m$ and $v$ amounts to enqueuing the closures in $T_m(v)$, i.e.,
\[
Q_{P-1} = \mbox{\tt enQueue}(Q_{P-1},f) \:\: \forall f \in T_m(v).
\]
\end{definition}

\noindent
Finally, the protocol provides two methods for removing and inserting
triggers (an opaque protocol) directly. The \texttt{CPTrigger}
protocol encapsulates a closure and includes data structure to remove
them in constant time. Section~\ref{sec:fdservice} illustrates the use
of triggers for implementing a propagator for $\sum_{i=0}^n b_i \geq
c$.

\subsection{Informers}

In addition to messaging via propagation, the \ocp{} microkernel
offers a simple messaging abstraction for thread-aware multicasting
through the \texttt{ORInformer} abstract data type. A similar idea was
already present in \comet{} where ``events'' were used for decoupling
meta-strategies in CBLS~\cite{CometBook} supporting parallel
search~\cite{COR09,IJOC09} and implementing
visualizations~\cite{Dooms:2009}.  \ocp{} generalizes it further.  An
\texttt{ORInformer} embodies the idea of the publish-subscribe design
pattern~\cite{DPGof94} and extends it to a concurrent setting. Two
protocols are shown in Figure~\ref{fig:orinformer}
\begin{figure}[tb]
\begin{lstlisting}
@protocol ORInformer<NSObject>
-(void) whenNotifiedDo: (id) func;
-(void) wheneverNotifiedDo: (id) func;
@end

@protocol ORIdInformer<ORInformer>
-(void) notifyWith: (id) a0;
@end
\end{lstlisting}
\vspace{-4mm}
\caption{The \texttt{ORInformer} Messaging Abstraction.}
\label{fig:orinformer}
\end{figure}
and specify the abstract informer protocol and a concrete informer
protocol.  Informally speaking, the abstract protocol in lines 1--4
provides two methods \texttt{whenNotifiedDo:} and
\texttt{wheneverNotifiedDo:} receive two first-order functions to be
executed only when (respectively each time) the informer is
notified. The facility is convenient to request the execution of an
arbitrary piece of code when some notification occurs.
The concrete protocol in lines 6--8 extends the core capability with a
single notification method \texttt{notifyWith:} responsible for
relaying its argument to every subscriber. An informer implementation
maintains two lists of first-order functions (\texttt{once} and
\texttt{always}). When an occurrence is notified via the notification
API, the closures are scheduled for execution \textit{in the thread
  that performed the subscription}.

\subsection{Microkernels as First-Class Objects}

Microkernels in \ocp{} are first-class objects and can have their own
dedicated propagation algorithms by overloading the {\tt propagate}
method. Each microkernel, except the root microkernel, has a parent
microkernel that initiates its propagation. All the events presented
earlier can be generalized to specify the microkernel in which they
must dispatched. Consider, for instance, a closure event of the form
$\langle f, p, k \rangle$, where $k$ is a microkernel. Dispatching
such an event consists of two steps: (1) Scheduling the event, i.e.,
\[
Q_{p}^k = \mbox{\tt enQueue}(Q_p^k,f),
\]
where $Q_p^g$ denotes the queue of priority $p$ in microkernel $k$; and (2)
Dispatching the propagation of microkernel $k$, i.e., 
\[
Q_{p_k}^{u_k} = \mbox{\tt enQueue}(Q_{p_k}^{u_k},\lambda.{\tt propagate}(k))
\]
where $p_k$ is the priority of microkernel $k$ and $u_k$ is its parent
microkernel. Groups \cite{LagerkvistS09} can be naturally implemented
as microkernels in \ocp{} by overloading method {\tt propagate}. 


\section{A Finite Domain Service}
\label{sec:fdservice}
This section shows how to create a finite-domain solver over integers
o top of the microkernel. It reviews some of the core ideas underlying
the \ocp{} implementation.

\subsection{Variables}

\begin{definition}[Variable]
  A variable is a tuple $\langle D,m,M,B,I,L,T\rangle $ associating a
  domain representation with five constraint event lists and a trigger
  map. The event lists $m, M, B, I, L$ are associated with with the
  constraint events monitoring changes to the minimum $m$, changes to
  the maximum $M$, changes to either bounds $B$, instantiation $I$,
  loss of value $L$. The trigger map $T$ monitors value losses as
  well.
\end{definition}

\begin{figure}[t]
\begin{lstlisting}
@interface CPIntVar {
	id<CPEngine> _engine;
	id<CPDom> _dom;
	id<CPClosureEvent>  _min,_max,_bounds,_bind;
	id<CPValueEvent> _loss;
	id<CPTriggerMap> _triggers;
}
-(CPIntVar*) initVar: (CPEngine*) engine low: (ORInt) low  up: (ORInt) up;
-(void) whenChangeMinDo: (void^()) f priority: (ORInt) p;
-(void) whenChangeMaxDo: (void^())f priority: (ORInt) p;
-(void) whenChangeBoundsDo: (void^())f priority: (ORInt) p;
-(void) whenLoseValueDo: (void^(ORInt)) f;
-(void) whenLoseValue: (ORInt) v trigger: (void^()) f;
-(void) updateMin: (ORInt) newMin;
-(void) updateMax: (ORInt) newMin;
-(void) removeValue: (ORInt) value;
@end
\end{lstlisting}
\caption{An Integer Variable Definition.}
\label{fig:intvar}
\end{figure}
\noindent Variables are a key building block for stating constraints
and \ocp{} provides (a superset) of the class definition shown in
Figure~\ref{fig:intvar}.  It includes the APIs needed to register
different types of events. The {\tt \_dom} instance variable is a
reference to a suitable domain representation such as a range, a
bit-vector, or a list of intervals. Methods such as
\texttt{whenChangeMinDo:} and \texttt{whenLoseValueDo:} simply
delegate to their respective event lists: Their implementation
is as follows:
\begin{lstlisting}
-(void) whenChangeMinDo:(void^())f priority:(ORInt)p {
	[_min insert:f withPriority:p];
}
-(void) whenLoseValueDo:(void^(ORInt))f  {
	[_loss insert:f];
}
-(void) whenLoseValue:(ORInt)v trigger:(void^())f {
	[_triggers loseTrigger:f forValue:v];
}
\end{lstlisting}
The methods responsible for domain updates are expected to schedule
the proper events. Consider method \texttt{removeValue:}
\begin{lstlisting}
-(void) removeValue: (ORInt) value {
   BOOL rMin = value == [_dom min];
   BOOL rMax = value == [_dom max];
   BOOL changed = [_dom remove:value];
   if (changed) {
      if (rMin) [_engine scheduleClosureEvt:_min];
      if (rMax) [_engine scheduleClosureEvt:_max];
      if (rMin || rMax) [_engine scheduleClosureEvt:_bounds];
      if ([_dom size] == 1) [_engine scheduleClosureEvt:_bind];
      [_engine scheduleValueEvt: _loss with: value];
      [_engine dispatch: _triggers with: value];
   }
}
\end{lstlisting}
The method performs the domain update but most of its body is devoted
to scheduling the relevant events. For instance, if the value removed
is the smallest value in the domain, it schedules the events in {\tt
  \_min}. If the domain is now a singleton, line 9 also schedules the
events on the {\tt \_bind} list. Finally, lines 10-11 schedule the
value events and dispatches the triggers.  If the domain update
results in a wipe-out, method {\tt remove:} method on the domain calls
the {\tt fail} method described in Figure~\ref{fun:tryfail} to report
the failure: There is no need to obfuscate the code of method {\tt
  removeValue:} with consistency tests.

Overall, the variable tracks response behaviors that are suitable for
each type of events and schedules the messages with the microkernel
whenever an event of that class is recognized. The number and the
semantics of the events are solely the variable responsibility and
completely orthogonal to the microkernel.

\subsection{Constraints}

This section reviews how to implement constraints using the
microkernel functionalities.

\paragraph{Closure-Based Propagation}

Consider constraint $x = y + c$ and its implementation in
Figure~\ref{eqbc} which is exclusively in method \texttt{post}.  When
posted, the constraint first updates the domains of $x$ and $y$.  It
then states (lines 11--13) that, whenever the lower or upper bound of
$x$ change, the specified closure should be executed. The process is
repeated for $y$ to update $x$.  The two closures capture \textit{all
  the names that are in the lexical scope of their
  definitions}. Namely, both closures capture the names {\tt \_x},
{\tt \_y}, {\tt \_c}, and {\tt self} and refer to them within their
implementations.  While seemingly innocuous, this capability is
essential to pass information to, and share information with,
the closures.

\begin{figure}[t]
\begin{lstlisting}
@implementation CPEqualBC
-(id) init: (CPIntVar*) x equalTo: (CPIntVar*) y  plus: (ORInt) c {
   self = [super initCPCoreConstraint: [x engine]];
   _x = x;_y = y;_c = c;
   return self;
}
-(void) post {
   [_y updateMin: _x.Min - _c andMax: _x.max - _c];
   [_x updateMin: _y.min + _c andMax: _y.max + _c];
   if (!_x.bound)
       [_x whenChangeBoundsDo: ^ {
       	  [_y updateMin:_x.min - _c andMax:_x.max - _c];       	  
       }];
   if (!_y.bound)
       [_y whenChangeBoundsDo: ^ {
       	  [_x updateMin:_y.min + _c andMax:_y.max + _c];
       }];
 }
@end
\end{lstlisting}
\vspace{-4mm}
\caption{A Bound-Consistency Propagator for $x= y+c$.}
\label{eqbc}
\end{figure}

\paragraph{Value-Based Propagation}

\begin{figure}[t]
\begin{lstlisting}
@implementation CPEqualDC
-(id) init: (CPIntVar*) x equalTo: (CPIntVar*) y  plus: (ORInt) c {
   self = [super initCPCoreConstraint:[x kernel]];
   _x = x;_y = y;_c = c;
   return self;
}
-(void) post {
   if (bound(_x)) 
      [_y bind: _x.min - _c];
   else if (bound(_y)) 
      [_x bind: _y.min + _c];
   else {
      [_x updateMin: _y.min + _c andMax: _y.max + _c];
      [_y updateMin: _x.min - _c andMax: _x.max - _c];
      for(ORInt i = _x.min;i <= _x.max; i++)
         if (![_x member: i])
            [_y remove: i - _c];
      for(ORInt i = _y.min; i <= _y.max; i++)
         if (![_y member: i])
            [_x remove: i + _c];

      [_x whenLoseValueDo: ^(ORInt v) { [_y remove: v - _c];}];
      [_y whenLoseValueDo: ^(ORInt v) { [_x remove: v + _c];}];
      [_x whenBindDo: ^{ [_y bind: _x.min - _c];}];
      [_y whenBindDo: ^{ [_x bind: _y.min + _c];}];
   }
}
\end{lstlisting}
\vspace{-4mm}
\caption{A Domain Consistency Propagator for $x= y + c$.}
\label{eqdc}
\end{figure}

Consider a domain-consistent propagator for the same constraint.
Figure~\ref{eqdc} showns the bulk of the propagator implementation
which also takes place in the \texttt{post} method. Lines 8--12 cover
the trivial cases where one of the variables is bound: The other
variable is simply updated accordingly. Lines 13--14 initiate the
domain filtering of $x$ and $y$ by tightening their respective
bounds. Lines 15--20 proceed with two tight loops to discard the
images of values that are not in the domains. Lines 22--23 setup two
closures to respond to value losses in the domains. The
implementations take constant time and simply remove the correct image
from the domain of the other variable. Lines 24--25 are handling the
constraint events that arise when a variable is bound. 

The code is simple thanks to the use of closures which blend
references to parameters (e.g., $v$) and to local and instance
variables. The code mimics the inference rules and lexically binds the
specifications of events (\texttt{whenXXX} messages sent to variables)
with the proper response (the closures passed to the message).

\paragraph{Trigger-Based Propagation}

\begin{figure}[t]
\begin{lstlisting}
@interface CPSumBoolGeq : CPCoreConstraint {
    id<CPIntVarArray> _x;
    ORInt             _n;
    ORInt             _c;
    id<CPTrigger>*   _at; // the c+1 triggers.
    ORInt*           _nt; // local identifiers of non-triggers
    ORInt          _last;
}
-(id) initCPSumBool: (id) x geq: (ORInt) c;
-(void) post;
@end
\end{lstlisting}
\vspace{-4mm}
\caption{Class definition for the Linear inequality propagator.}
\label{propag:def:geq}
\end{figure}

Consider the linear inequality constraint $\sum_{i=0}^n x_i \geq c$,
where each $x_i$ is a boolean variable and $c$ is a constant. The key
idea behind the implementation is to monitor the loss of the value
{\tt true} from the domains. Triggers are useful to listen to
\textit{only} $c+1$ variables among $\{b_0,\cdots,b_n\}$. As soon as a
variable loses its {\tt true} value, the constraint seeks another
witness among the variables not listened to. If no such witness can be
found, the remaining variables must necessarily be all true.

Figure~\ref{propag:def:geq} provides the class definition for the
propagator.  The class has a few attributes to track the input array
$x$, its size, the constant $c$, the array of triggers {\tt \_at} as
well as the array {\tt \_nt} of variables not listened to. Following
the original algorithm in \cite{Gent06}, the instance variable {\tt
  \_last} tracks the place where the implementation resumes its
scanning for another witness.

\begin{figure}
\begin{lstlisting}
@implementation CPSumBoolGeq
-(id) initCPSumBool: (id<CPIntVarArray>) x geq: (ORInt) c {
   _self = [super initCPCoreConstraint:[x[x.low] engine]];	
   _x = x;
   _n = [x count];
   _c = c;
   _at = 0;
   _id = 0;
   _nt = 0;
   return self;
}

-(void) post {
    _at = malloc(sizeof(id<CPTrigger>)*(_c+1));
    _id = malloc(sizeof(ORInt)*(_c + 1));
    _nt = malloc(sizeof(ORInt)*(_nb - _c - 1));
    int nbTrue = 0,nbPos  = 0;
    for(ORInt i=0;i < _n;i++) {
       [_x[i] updateMin:0 andMax:1];
       nbTrue += (_x[i].bound && _x[i].min == true);
       nbPos  += !_x[i].bound;
    }
    if (nbTrue >= _c) return;
    if (nbTrue + nbPos < _c)  fail();
    if (nbTrue + nbPos == _c) {
        for(ORInt i=0;i<_n;++i) {
           if (_x[i].bound) 
              continue;
           [_x[i] bind: true];
        }
        return;      
    }
    ORInt listen = _c+1;
    ORInt nbNW   = 0;     // number of variables not watched
    for(ORInt i=_n-1;i >= 0;--i) 
        if (listen > 0 && _x[i].max == true) { 
            --listen; 
            _id[listen] = i; 
            _at[listen] = [_x[i] whenLoseValue: true trigger: ^{
                   ORInt j = _last;
                   BOOL jOk = NO;
                   if (_last >= 0) { // seek alternate support
                      do {
                         j=(j+1) % (_n - _c - 1);
                         jOk = [_x[_nt[j]] member: true];
                      } 
                      while (j != _last && !jOk);
                   }
                   if (jOk) {
                       ORInt nextVar = _nt[j];
                       id<CPTrigger> toMove = _at[listen];
                       [_x[_id[listen]].triggers removeTrigger: toMove];
                       _nt[j] = _id[listen];
                       [_x[nextVar].triggers addTrigger: toMove];
                       _id[listen] = nextVar; 
                       _last = j;
                   } 
                   else {  // no alternative support. Bind all watched.
                       for(ORInt k=0;k<_c+1;k++)
                         if (k != listen) 
                           [_x[_id[k]] bind:true];                         
                   }
            }];                           
        } 
        else 
           _nt[nbNW++] = i;
    _last = _n - _c - 2; 
}
@end
\end{lstlisting}
\vspace{-4mm}
\caption{Linear Inequality propagator.}
\label{propag:geq}
\end{figure}

Figure~\ref{propag:geq} shows the \textit{entire} implementation of
the propagator. The constructor in lines 2--10 is straightforward. The
{\tt post} method first allocates memory to hold the triggers, the
identification of the variables they are listening to, as well as the
identifiers of the variables not monitored. Lines 18--22 ensure that
each variable is boolean and compute the number of variables already
bound to true. Lines 23--31 deal with the trivial cases when the
constraint is obviously true, always false, or just satisfiable if all
possible variable are bound to true now.
The loop spanning lines 36--66 is the core of the implementation. It
looks for $c+1$ variables that still have {\tt true} in their domains.
Each time such a variable is found, a trigger is added to the trigger
map and recorded in array {\tt \_at} (line 39). Line 38 also remembers
that the trigger listens to variable {\tt i} at this point.  Line 66
stores in {\tt \_nt} the variables that are not listened to (because
they no longer have true in their domains or because $c+1$ variables
are already listened to.).

The trigger listens to the loss of the {\tt true} value. If the
variable loses this value, the closure in lines 40--63) is executed.
As usual, each closure captures the local variables in scope and
remembers their values at the time of the closure creation. In
particular, the variable {\tt listen} always correctly refers to the
right entry in {\tt \_at}. The closure accomplishes the following
tasks.  First, it seeks a replacement witness among the variables in
{\tt \_nt} (lines 42-- 48).  If such an alternate support is found,
the trigger is moved to the new supporting variable (lines 50--56). If
no such support is found, the remaining $c$ variables with triggers
must necessarily be bound to {\tt true}, which is done by the code in
lines 59--61.

\paragraph{Constraint-Based Propagation}

Many constraints are implemented through two methods: a {\tt post}
method that initializes some data structures and possibly create some
events to update them dynamically; and a {\tt propagate} method that
performs the domain reduction based on these data structures. In
particular, this is the case of many global constraints. For illustration
purposes, Figure~\ref{eqbc-c} depicts a constraint-based propagation
of constraint $x = y + c$. Observe line 10 where the closure simply calls 
method {\tt propagate}. This pattern is so frequent that it is encapsulated
in methods of the form 
\begin{lstlisting}
-(void) whenChangeMinPropagate: (CPConstraint) c priority: (ORInt) p
\end{lstlisting}
in the API of the variables and the kernel. It is also optimized to avoid 
redundant calls to {\tt propagate}. 

\begin{figure}[t]
\begin{lstlisting}
@implementation CPEqualBC
-(id) init: (CPIntVar*) x equalTo: (CPIntVar*) y  plus: (ORInt) c {
   self = [super initCPCoreConstraint: [x engine]];
   _x = x;_y = y;_c = c;
   return self;
}
-(void) post {
   [self propagate];
   if (!_x.bound)
       [_x whenChangeBoundsDo: ^ { [self propagate]; }];
   if (!_y.bound)
       [_y whenChangeBoundsDo: ^ { [self propagate]; }];
 }
-(void) propagate {
   [_x updateMin:_y.min + _c andMax:_y.max + _c];
   [_y updateMin:_x.min - _c andMax:_x.max - _c];       	  
 }
@end
\end{lstlisting}
\vspace{-4mm}
\caption{A Constraint-Based Propagator for $x= y+c$.}
\label{eqbc-c}
\end{figure}

\paragraph{Flexibility of the Microkernel}

It is useful to conclude this section by highlighting the flexibility
of the microkernel on a slightly more complicated propagator. Consider
the element constraint $z = y[x]$, where $x$ and $z$ are variables and
$y$ is an array of variables. Assume, in addition, that the implementation
maintains the following data structures:
\begin{itemize}
\item $I_k = D(z) \cap D(y_k)$: The intersection between the $k^{th}$ entry of array $y$ and $z$;
\item $H$: A local copy of the domain of variable $x$;
\item $s_v = |\{ k \in D(x) | v \in D(y_k)\}|$: The number of supports
for value $v$ of $z$.
\end{itemize}
An implementation enforcing domain consistency may perform the
following actions when variable $y_k$ loses value $k \in H$:
\[
\left\{\begin{array}{l}
s_v \leftarrow s_v -1 \\
I_k \leftarrow I_k \setminus \{v\} \\
I_k = \emptyset \Rightarrow D(x) \leftarrow D(x) \setminus \{k\}\\
s_v = 0 \Rightarrow D(z) \leftarrow D(z) \setminus \{v\}. \\
\end{array}\right.
\]

\begin{figure}[tb]
\begin{lstlisting}
@implementation CPElementVarDC
-(void) post {
    ...                                        // Setup the data structures.
    for(int k=_x.min;k <= _x.max;k++) {        // Rules from $y_k$ to x and z
      if ([_x member:k] && !y[k].bound) {
         [y[k] whenLoseValue: ^(ORInt v) {
            if ([_H get:k]) {
               [_s[v] decr];
               [_I[k] set: v at: false];
               if ([_I[k] empty]) 
                  [_x removeValue: k];;               
               if ([_s[v] value] == 0)
                  [_z removeValue: v];
            }
         }];
      }
   }
}
@end
\end{lstlisting}
\vspace{-4mm}
\caption{The Domain-Consistent Element Propagator.}
\label{element}
\end{figure}

Figure~\ref{element} describes how to implement these ideas in
\ocp{}. All the actions are enclosed in a first-order function that
uses both the local variables in scope and the removed value, which is an
argument to the first-order function. It is a compact implementation
where the event and its response are jointly specified.

A similar behavior can be achieved in \comet{} using method {\tt
  valRemoveIdx} on line 9 of the {\tt AC5Constraint} protocol in
Figure~\ref{cometapi}. Method {\tt valRemoveIdx} was added to the
protocol to implement such propagation rule: Indeeed it is necessary
to transmit the index $k$ of the variable $y_k$ to achieve the desired
behavior. Hencem, while no extension to the \ocp{} microkernel were
necessary to implement this constraint, the \comet{} API had to be
duplicated to integrate the rather ad-hoc concept of index.  Moreover,
the \comet{} code loses the textual proximity between the event and
its response.

\subsection{Implementating Black-Box Searches}

This section illustrates how to implement search heuristics using
informers. Consider, for instance, \textsc{Ibs}~\cite{Refalo04}.  The
heuristic requires that, after a branching decision $x=v$, the impact
of the assignment be evaluated and recorded in a data structure for
the heuristic to use during the next variable selection. The impact
depends upon the outcome of the propagation. When $x=v$ succeeds, the
impact $I(x=v) = 1.0 - \frac{{\cal S}(P^k)}{{\cal S}(P^{k-1})}$ where
${\cal S}(P)$ evaluates an upper-bound on the size of the search space
$P$ that uses the product of the domain sizes. Note that $P^{k-1}$ and
$P^{k}$ respectively refers to the state before and after enforcing
$x=v$.  When $x=v$ fails, the impact is maximal, i.e., $I(x=v)=1.0$.

To implement this logic, two informers relay the outcome of posting
the branching decision. Consider the code in
Figure~\ref{ex:inform1}. The concrete CP solver holds two informer
instances {\tt \_returnLabel} and {\tt \_failLabel}. The labeling
method of the concrete solver uses (line 8) the {\tt enforce:} method
of the microkernel to propagate the effects of $x=v$. If the outcome
is a failure, line 10 notifies the {\tt \_failLabel} informer and
proceeds by asking the explorer to backtrack on line 11. If $x=v$
succeeds, the control flows to line 13 where the solver notifies the
{\tt \_returnLabel} informer.

\begin{figure}[t]
\begin{lstlisting}
@implementation CPSolver  {
	id<ORInformer> _returnLabel;
	id<ORInformer> _failLabel;
}
...
-(void) label: (id<CPIntVar>) var with: (ORInt) val
{
   ORStatus status = [_engine enforce: ^ {[var bind: val];}];
   if (status == ORFailure) {
      [_failLabel notifyWith:var andInt:val];
      [_search fail];
   }
   [_returnLabel notifyWith:var andInt:val];
}
@end
\end{lstlisting}
\caption{Notifying labeling outcomes.}
\label{ex:inform1}
\end{figure}

The object encaspulating the \textsc{Ibs} implementation can subscribe
to both notifications and execute code fragments to compute the actual
impact. An abridged version of \textsc{Ibs} is shown in
Figure~\ref{ex:ibs}.  The {\tt initInternal} method receives the array
of variables. Line 5 creates a monitor daemon and attaches it to every
variables. This monitor is responsible for computing $\frac{{\cal
    S}(P^k)}{{\cal S}(P^{k-1})}$ automatically with an amount of work
linear in the number of variables affected by $x=v$.  Lines 9--11
setup a listener on the solver's {\tt retLabel} informer. The
listening closure computes the actual impact from the search space
reduction established by the monitor and updates the {\tt \_impacts}
dictionary accordingly. Lines 12--14 echo the same logic when $x=v$
fails with a listener on the {\tt failLabel} informer of the
solver. The result is a nice modular implementation of \textsc{IBS}.

\begin{figure}[t]
\begin{lstlisting}
@implementation CPIBS 
-(void) initInternal: (id<ORVarArray>) t
{
   _vars = t;
   _mon = [[CPStatisticsMonitor alloc] initCPMonitor:[_cp engine] vars: t];
   [_engine add:_mon];
   ...
   [self initImpacts];       
   [[_cp  retLabel] wheneverNotifiedDo:^void(id var,ORInt val) {
     [[_impacts forVar:var.getId andVal:val] addImpact:1.0-[_mon reduction]];
   }];
   [[_cp failLabel] wheneverNotifiedDo:^void(id var,ORInt val) {
     [[_impacts forVar:var.getId andVal:val] addImpact:1.0];
   }];
}
@end
\end{lstlisting}
\caption{An Abridged Implementation of \textsc{Ibs}.}
\label{ex:ibs}
\end{figure}

\section{Empirical Results}
\label{sec:empirical}

To measure the performance of \ocp{} microkernel, this section
compares its behavior (space and time performance) against the COMET
2.1.0 implementation. In particular, it reports on three sets of
experiments.
First, it considers micro-benchmarks where the bulk of the computation
time takes place inside the propagation engine due to a large number
of propagation events (propagators are fast and there is virtually no
search).
Second, it reports profiling benchmarks obtained from development tools
(i.e., \texttt{dtrace}) that measure the cost of each method and
functions in the implementation. This sheds some light on the cost of
dynamic dispatching.
Third, it selects representative application benchmarks featuring a
mix of global constraints, arithmetic constraints, reified
constraints, and logical constraints. In this case, there is an actual
effort expanded in the search, but the benchmarks offer some insights
about the cost of the propagation engine when embedded inside a real
solver and in realistic conditions.

\subsection{Micro-Benchmarks}

The micro-benchmarks represent the worst situation for the engine as
there are many events, each of which propagates quickly. These
benchmarks thus indicate the cost of generality and compositionality
in the microkernel.  Four models were considered:
\begin{description}

\item[\texttt{order}] correspond to a pathological model with $n$
  variables with a domain $1..n$ and $n-1$ binary constraints of the
  form $x_i < x_{i+1} \forall i \in 1..n-1$. Without a dedicated group
  and a custom scheduler exploiting Berge acyclicity, the propagation
  engine triggers a quadratic number of propagation events taking
  constant time.

\item[\texttt{magic/s}] is the magic series benchmark where each
  term $s_i$ is subjected to a counting constraints expressed
  algebraically as $s_i = \sum_{j \in 1..n} (s_j = i)$. There are no
  redundant constraints, the labeling is static, and the model
  searches for all solutions.

\item[\texttt{magic/r}] is the magic series again, but with
  the two traditional redundant constraints and a labeling procedure
  that considers the variables in a static order and chooses values in
  decreasing order.

\item[\texttt{slow}] is a benchmark used in the MiniZinc
  challenge ({\tt slowConvergence} and designed to ``stress test'' 
  propagation engines.

\end{description}

\begin{figure}[t]
\begin{center}
\includegraphics[width=0.48\textwidth]{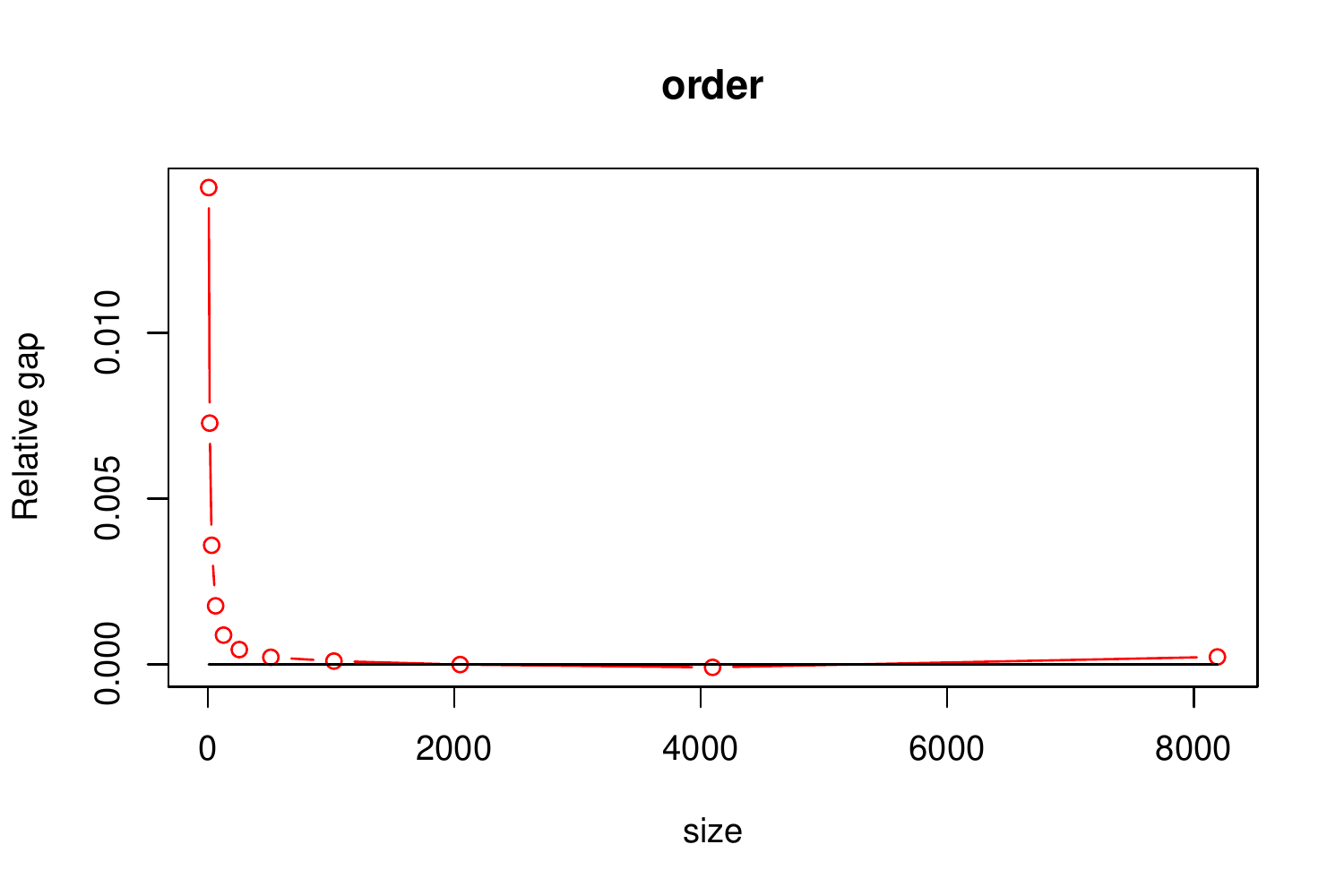}
\includegraphics[width=0.48\textwidth]{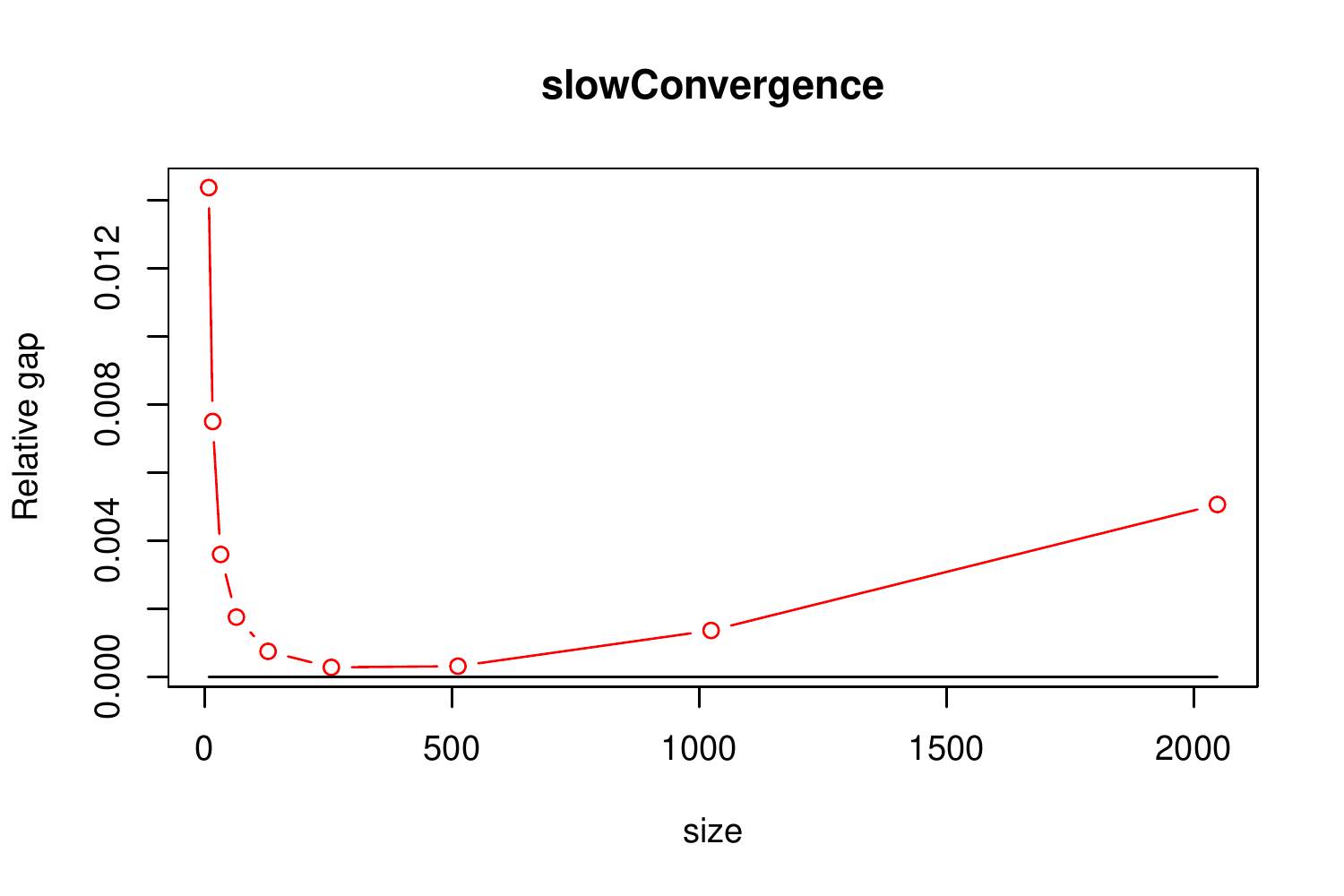}
\vspace{-4mm}
\includegraphics[width=0.48\textwidth]{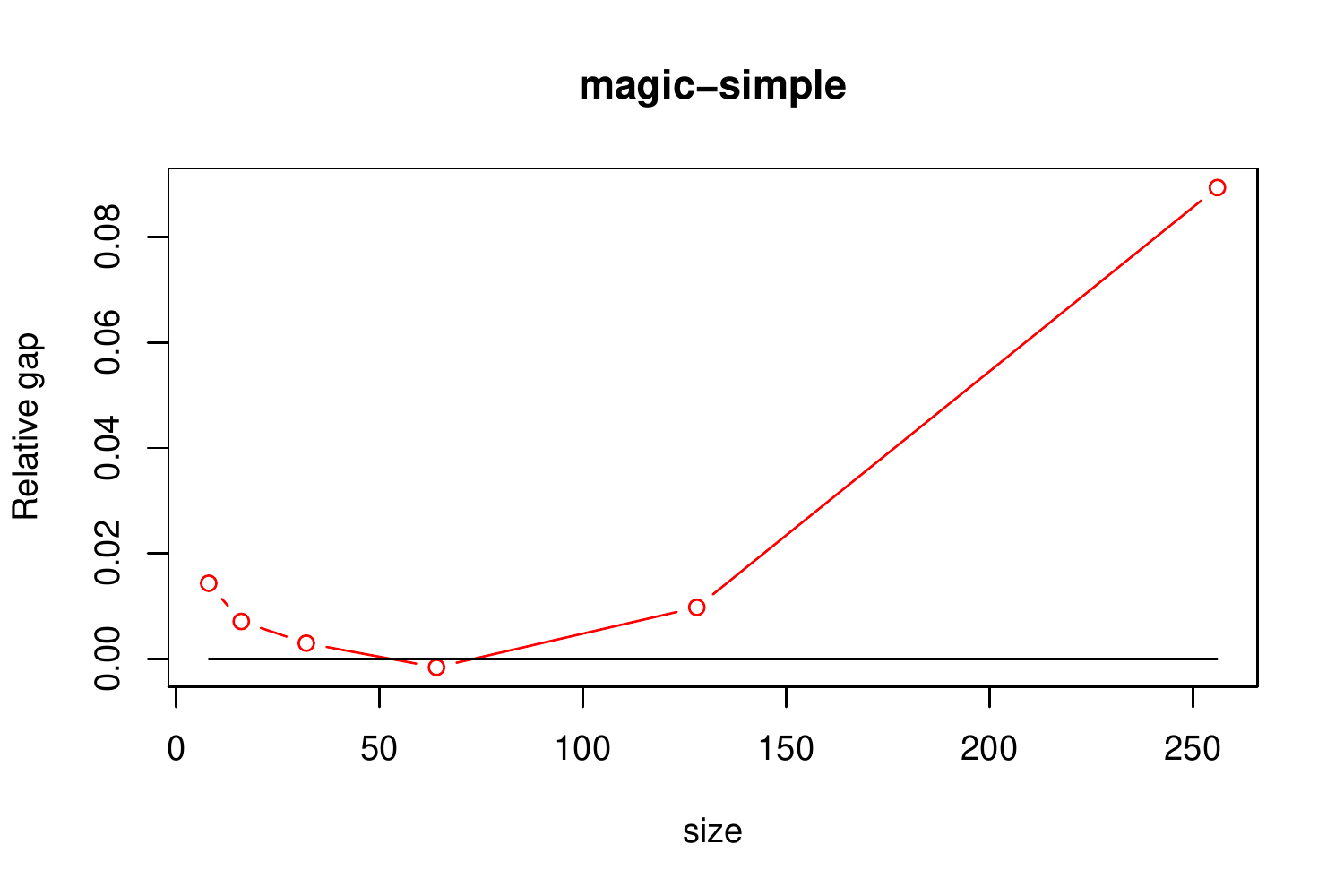}
\includegraphics[width=0.48\textwidth]{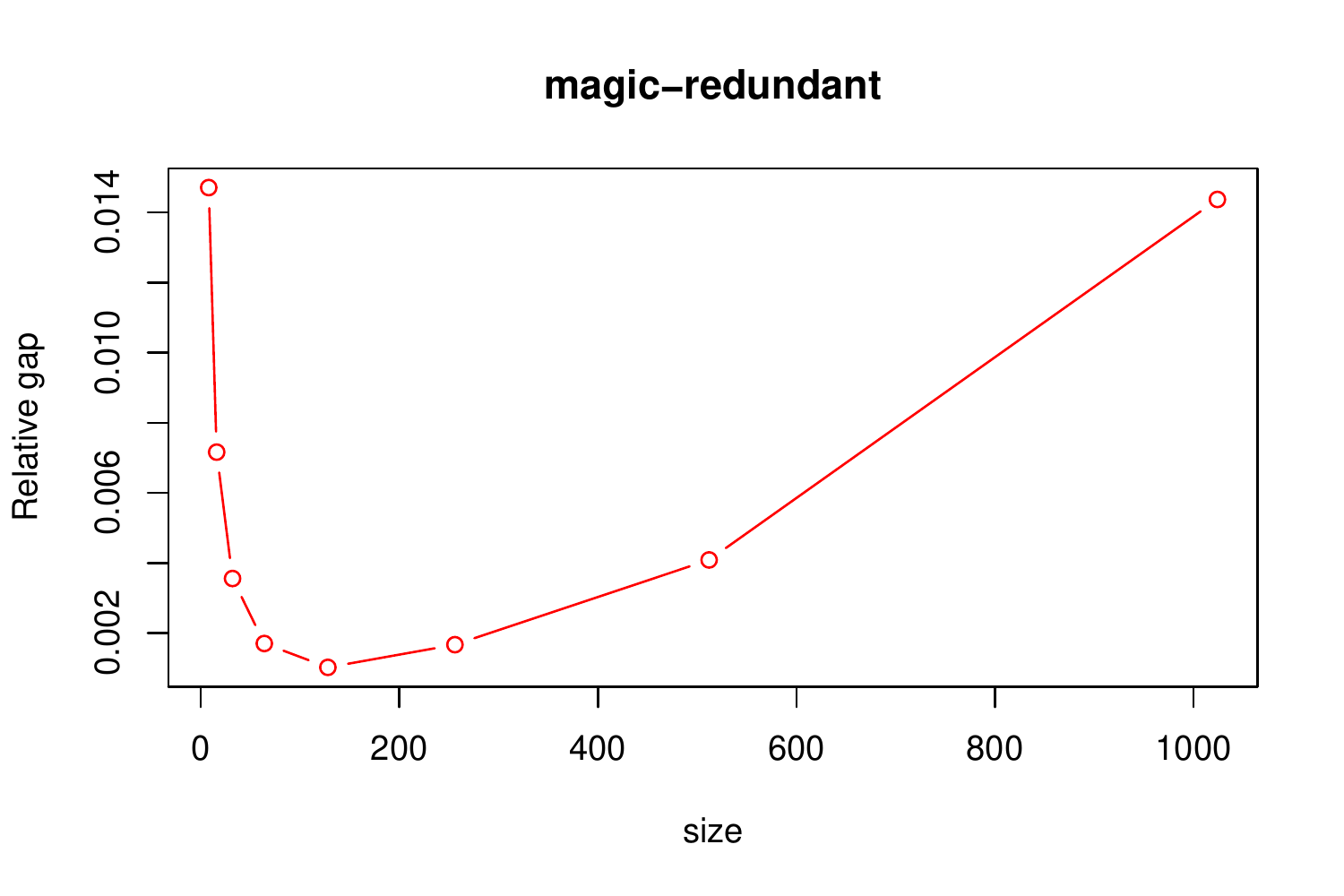}\\
\end{center}
\vspace{-2mm}
\caption{Comparative Performance of \comet{} and \ocp{} on
  Micro-benchmarks. The curve indicates the relative speed difference
  between \comet{} and \ocp{} scaled by size. A positive value
  indicates that \comet{} is slower than \ocp{}.}
\label{fig:micro}
\end{figure}

\noindent
In all cases, care was taken to make sure that the number of choices
made during the search were identical. Figure~\ref{fig:micro} offers a
quick overview of the comparative performance between the latest
version of \comet{} and \ocp{}. In particular, the curve reports the
relative speed difference as
\[
\frac{\mu(T_{\comet}) - \mu(T_{\ocp})}{size}
\]
Namely, a positive value indicates that \comet{} is slower and the
time difference is weighted by the benchmark size. Given that the
micro-benchmarks are deterministic, only 10 runs are included (to
account for speed variation caused by dynamic frequency scaling of the
CPU and/or activity of the operating system). 

The simplest micro-benchmark is \texttt{order} where the propagation
of each event runs in constant time and the volume of events grows
quadratically with the instance size. The initial, large advantage of
\ocp{} over \comet{} can be attributed to the fact that \ocp{} being
compiled while \comet{} use a Just-In-Time (JIT) compiler. However, as
the problem size grows to 8192, the two solvers are in a near tie with
a very minimal advantage to \ocp{}. It is worth noting that the
implementation of variables and domains in \ocp{} relies on dynamic
dispatching for the delivery of key messages such as updating the
bounds, whereas \comet{} uses a polymorphic
implementation in \texttt{C++}.
Benchmark \texttt{slowConvergence} is a slightly refined version of
\texttt{order}. One notices the same initial edge of \ocp{} over
\comet{} due to JIT phase. As instance size grows, the advantage
returns to \ocp{}. It is important to realize that the model is a
collection of many algebraic constraints and most of the runtime is
spent building the model which exercises the dynamic dispatching code
quite a lot.  Since the model creates a quadratic number of
constraints, its memory consumption is also a key factor. Note how,
for instances of size 2048, the model allocates up to 1.6 gigabytes of
memory (\comet{} uses 2 gigabytes for the same instance). The \ocp
model is shown in Figure~\ref{fig:slow} for completeness.
Benchmark \texttt{magic-simple} creates a large number of reified equalities and
auxiliary boolean variables. Instances of size $8..128$ were used in
the experiment and \ocp{} is almost always faster (with the exception
of instance 64). It is worth noting that the \ocp{} implementation
uses non-injective views for the reifications~\cite{DomainViews13}.
Benchmark \texttt{magic-redundant} has been evaluated for instances of
sizes $8..1024$. The curve shows the same shape as
\texttt{slowConvergence}.

\begin{figure}[tb]
\lstset{language=[Objective]C,mathescape=true,numbers=left,numbersep=1mm,xleftmargin=6pt,frame=lines,
tabsize=2,
basicstyle=\scriptsize\ttfamily,
morekeywords={for}
}
\begin{lstlisting}
id<ORModel> model = [ORFactory createModel];
ORInt n = [args size];
id<ORIntRange> dom = RANGE(model,0,10*n);
id<ORIntRange> R   = RANGE(model,0,n);
id<ORIntVarArray> y = [ORFactory intVarArray:model range:R domain:dom];
id<ORIntVarArray> x = [ORFactory intVarArray:model range:R domain:dom];
for(ORInt i=2;i<=n;i++)
   [model add:[[y[i-1] sub:y[i]] leq:@0]];         
for(ORInt i=1;i<=n;i++)
   [model add:[[y[0] sub:y[i]] leq:@(n-i+1)]];        
[model add:[[y[n] sub:x[0]] leq:@0]];        
for(ORInt i=1;i<=n-1;i++) 
   for(ORInt j=i+1;j<=n;j++) 
     [model add:[[x[i] sub:x[j]] leq:@0]];
[model add:[y[0] geq:@(n)]];
\end{lstlisting}
\vspace{-4mm}
\caption{The \ocp{} model for {\tt slowConvergence}.} 
\label{fig:slow}
\end{figure}

\begin{table}[tb]
{\scriptsize
\begin{center}
\noindent\pgfplotstabletypeset[
  columns={solver,name,size,nfail,nprop,cpu,sigcpu,mpeak},
  string replace*={ocp}{\ocp},
  string replace*={comet}{\comet},
  string replace*={magic-simple}{magic/s},
  string replace*={magic-redundant}{magic/r},
  string replace*={slowConvergence}{slow},
  columns/solver/.style ={string type,column name=$Solver$,column type=l|},
  columns/name/.style ={string type,column name=$Bench$,column type=l|},
  columns/size/.style ={string type,column name=$Size$,column type=r|},
  columns/cpu/.style={column name=$\mu(T_{cpu})$,column type=r,precision=2,fixed},
  columns/sigcpu/.style={column name=$\sigma(T_{cpu})$,column type=r,precision=2,fixed},
  columns/nfail/.style={column name=$Fail$,column type=r|,precision=1,fixed},
  columns/nprop/.style={column name=$Prop$,column type=r|,precision=1,fixed},
  columns/mpeak/.style={column name=$Peak (Kb.)$,column type=r,precision=1,fixed},
  col sep=comma,
  skip rows between index={34}{88},
  every head row/.style={before row=\toprule%
  ,after row=\midrule},
  every last row/.style={after row=\bottomrule}
]{
,solver,name,size,nfail,nprop,cpu,sigcpu,mpeak
1,ocp,magic-simple,8,19,669,0.0062,0.000918936583472681,516.5390625
2,ocp,magic-simple,16,51,7704,0.0124,0.00195505043981536,639.5359375
3,ocp,magic-simple,32,115,68351,0.0667,0.00411096095821889,1134.0015625
4,ocp,magic-simple,64,243,570197,0.754,0.00847217668475923,3281.2546875
5,ocp,magic-simple,128,499,4651609,12.5969,0.188384624284114,12027.1875
6,ocp,magic-simple,256,1011,37577124,300.2065,3.88695834382962,45662.5546875
7,ocp,magic-redundant,8,3,81,0.0053,0.000483045891539648,528.984375
8,ocp,magic-redundant,16,3,129,0.0074,0.000699205898780101,659.803125
9,ocp,magic-redundant,32,3,225,0.0132,0.000421637021355784,1171.0734375
10,ocp,magic-redundant,64,3,417,0.0371,0.00202484567313166,3372.1421875
11,ocp,magic-redundant,128,3,801,0.1284,0.00350238014308366,12186.4828125
12,ocp,magic-redundant,256,3,1569,0.495,0.00852447456836296,46016.6234375
13,ocp,magic-redundant,512,3,3105,2.0968,0.0341298175272655,180787.3015625
14,ocp,magic-redundant,1024,3,6177,8.2993,0.121786744398194,720021.3859375
15,ocp,slowConvergence,8,0,59,0.006,0,481.2359375
16,ocp,slowConvergence,16,0,183,0.007,0,562.3203125
17,ocp,slowConvergence,32,0,623,0.013,0.00163299316185545,839.7875
18,ocp,slowConvergence,64,0,2271,0.0319,0.001523883926755,2056.7796875
19,ocp,slowConvergence,128,0,8639,0.1096,0.00411501316320292,7114.215625
20,ocp,slowConvergence,256,0,33663,0.4226,0.0124650622853549,27250.7578125
21,ocp,slowConvergence,512,0,132863,1.7826,0.0295604841337591,102743.834375
22,ocp,slowConvergence,1024,0,527871,7.0818,0.165758056616664,408484.7453125
23,ocp,slowConvergence,2048,0,2104319,28.6943,0.698442242899631,1632267.0203125
24,ocp,order,8,0,28,0.0069,0.000875595035770913,440.303125
25,ocp,order,16,0,120,0.0076,0.000699205898780101,449.984375
26,ocp,order,32,0,496,0.0072,0.000632455532033676,464.5359375
27,ocp,order,64,0,2016,0.0084,0.00126491106406735,499.0609375
28,ocp,order,128,0,8128,0.0099,0.00128668393770792,567.1046875
29,ocp,order,256,0,32640,0.0165,0.00190029237516523,727.7859375
30,ocp,order,512,0,130816,0.0342,0.00229975844142138,1017.265625
31,ocp,order,1024,0,523776,0.0957,0.00527151675411251,2035.3125
32,ocp,order,2048,0,2096128,0.348,0.0073181661333667,4005.5234375
33,ocp,order,4096,0,8386560,1.3162,0.152907234041501,9903.06875
34,ocp,order,8192,0,33550336,3.6949,0.324936728883913,35478.8234375
35,ocp,order2,8,8184,130956,0.0624,0.00171269767715535,20231.76875
36,ocp,order2,16,8176,261720,0.0742,0.00122927259430572,20235.2640625
37,ocp,order2,32,8160,522672,0.1012,0.0019888578520235,20242.6296875
38,ocp,order2,64,8128,1042272,0.1567,0.00283039062871384,20262.8734375
39,ocp,order2,128,8064,2072256,0.259,0.00326598632371091,20294.6359375
40,ocp,order2,256,7936,4095360,0.4694,0.00699523647443981,20380.684375
41,ocp,order2,512,7680,7994112,0.8847,0.00958065179874987,20504.7140625
42,ocp,order2,1024,7168,15201792,1.6376,0.0239499478078763,21136.7796875
43,ocp,order2,2048,6144,27257856,2.8916,0.022969545054267,21903.753125
44,ocp,order2,4096,4096,41932800,4.4515,0.140523821626244,23997.1890625
45,comet,magic-simple,8,19,1401,0.121,0.00316227766016838,32768
46,comet,magic-simple,16,51,15501,0.126,0.00516397779494323,32768
47,comet,magic-simple,32,115,131904,0.162,0.00421637021355784,32768
48,comet,magic-simple,64,243,1063720,0.651,0.0185292561462497,65536
49,comet,magic-simple,128,499,8490232,13.848,0.283619463365969,65536
50,comet,magic-simple,256,1011,67729560,323.097,0.807452096962344,131072
51,comet,magic-redundant,8,3,129,0.123,0.00483045891539648,32768
52,comet,magic-redundant,16,3,361,0.122,0.00421637021355784,32768
53,comet,magic-redundant,32,3,1209,0.127,0.00483045891539648,32768
54,comet,magic-redundant,64,3,4441,0.146,0.00516397779494321,65536
55,comet,magic-redundant,128,3,17049,0.259,0.0119721899973786,65536
56,comet,magic-redundant,256,3,66841,0.922,0.0113529242439509,131072
57,comet,magic-redundant,512,3,264729,4.189,0.0255821118057999,524288
58,comet,magic-redundant,1024,3,1053721,23.017,0.0352924291535099,2097152
59,comet,slowConvergence,8,0,103,0.121,0.00316227766016838,32768
60,comet,slowConvergence,16,0,335,0.127,0.0125166555703457,32768
61,comet,slowConvergence,32,0,1183,0.128,0.00421637021355784,32768
62,comet,slowConvergence,64,0,4415,0.144,0.006992058987801,32768
63,comet,slowConvergence,128,0,17023,0.205,0.00527046276694729,65536
64,comet,slowConvergence,256,0,66815,0.493,0.0163639169448448,65536
65,comet,slowConvergence,512,0,264703,1.941,0.0260128173535023,262144
66,comet,slowConvergence,1024,0,1053695,8.474,0.0330655913803659,524288
67,comet,slowConvergence,2048,0,4204543,39.059,0.350917844136393,2097152
68,comet,order,8,0,28,0.122,0.00421637021355784,32768
69,comet,order,16,0,120,0.124,0.0126491106406735,32768
70,comet,order,32,0,496,0.122,0.00421637021355784,32768
71,comet,order,64,0,2016,0.121,0.00316227766016838,32768
72,comet,order,128,0,8128,0.122,0.00421637021355784,32768
73,comet,order,256,0,32640,0.129,0.00316227766016838,32768
74,comet,order,512,0,130816,0.14,0,32768
75,comet,order,1024,0,523776,0.19,0.0176383420737639,32768
76,comet,order,2048,0,2096128,0.322,0.00421637021355784,65536
77,comet,order,4096,0,8386560,0.916,0.0201108041719978,65536
78,comet,order,8192,0,33550336,5.474,0.0340587727318529,65536
79,comet,order2,8,8184,130956,0.249,0.0296085573216033,65536
80,comet,order2,16,8176,261720,0.25,0.00816496580927727,65536
81,comet,order2,32,8160,522672,0.283,0.0211081869319834,65536
82,comet,order2,64,8128,1042272,0.342,0.0244040069569642,65536
83,comet,order2,128,8064,2072256,0.453,0.0194650684275419,65536
84,comet,order2,256,7936,4095360,0.67,0.0205480466765632,65536
85,comet,order2,512,7680,7994112,1.094,0.0422163738639668,65536
86,comet,order2,1024,7168,15201792,1.866,0.0386436713231718,65536
87,comet,order2,2048,6144,27257856,3.225,0.0683536555146996,65536
88,comet,order2,4096,4096,41932800,5.922,0.185998805253033,65536
}
\end{center}
}
\vspace{-2mm}
\caption{Micro-benchmarks (\ocp).}
\label{tab:micro1}
\vspace{-6mm}
\end{table}
\begin{table}[tbh]
{\scriptsize
\begin{center}
\noindent\pgfplotstabletypeset[
  columns={solver,name,size,nfail,nprop,cpu,sigcpu,mpeak},
  string replace*={ocp}{\ocp},
  string replace*={comet}{\comet},
  string replace*={magic-simple}{magic/s},
  string replace*={magic-redundant}{magic/r},
  string replace*={slowConvergence}{slow},
  columns/solver/.style ={string type,column name=$Solver$,column type=l|},
  columns/name/.style ={string type,column name=$Bench$,column type=l|},
  columns/size/.style ={string type,column name=$Size$,column type=r|},
  columns/cpu/.style={column name=$\mu(T_{cpu})$,column type=r,precision=2,fixed},
  columns/sigcpu/.style={column name=$\sigma(T_{cpu})$,column type=r,precision=2,fixed},
  columns/nfail/.style={column name=$Fail$,column type=r|,precision=1,fixed},
  columns/nprop/.style={column name=$Prop$,column type=r|,precision=1,fixed},
  columns/mpeak/.style={column name=$Peak (Kb.)$,column type=r,precision=1,fixed},
  col sep=comma,
  skip rows between index={0}{44},
  skip rows between index={78}{90},
  every head row/.style={before row=\toprule%
  ,after row=\midrule},
  every last row/.style={after row=\bottomrule}
]{
,solver,name,size,nfail,nprop,cpu,sigcpu,mpeak
1,ocp,magic-simple,8,19,669,0.0062,0.000918936583472681,516.5390625
2,ocp,magic-simple,16,51,7704,0.0124,0.00195505043981536,639.5359375
3,ocp,magic-simple,32,115,68351,0.0667,0.00411096095821889,1134.0015625
4,ocp,magic-simple,64,243,570197,0.754,0.00847217668475923,3281.2546875
5,ocp,magic-simple,128,499,4651609,12.5969,0.188384624284114,12027.1875
6,ocp,magic-simple,256,1011,37577124,300.2065,3.88695834382962,45662.5546875
7,ocp,magic-redundant,8,3,81,0.0053,0.000483045891539648,528.984375
8,ocp,magic-redundant,16,3,129,0.0074,0.000699205898780101,659.803125
9,ocp,magic-redundant,32,3,225,0.0132,0.000421637021355784,1171.0734375
10,ocp,magic-redundant,64,3,417,0.0371,0.00202484567313166,3372.1421875
11,ocp,magic-redundant,128,3,801,0.1284,0.00350238014308366,12186.4828125
12,ocp,magic-redundant,256,3,1569,0.495,0.00852447456836296,46016.6234375
13,ocp,magic-redundant,512,3,3105,2.0968,0.0341298175272655,180787.3015625
14,ocp,magic-redundant,1024,3,6177,8.2993,0.121786744398194,720021.3859375
15,ocp,slowConvergence,8,0,59,0.006,0,481.2359375
16,ocp,slowConvergence,16,0,183,0.007,0,562.3203125
17,ocp,slowConvergence,32,0,623,0.013,0.00163299316185545,839.7875
18,ocp,slowConvergence,64,0,2271,0.0319,0.001523883926755,2056.7796875
19,ocp,slowConvergence,128,0,8639,0.1096,0.00411501316320292,7114.215625
20,ocp,slowConvergence,256,0,33663,0.4226,0.0124650622853549,27250.7578125
21,ocp,slowConvergence,512,0,132863,1.7826,0.0295604841337591,102743.834375
22,ocp,slowConvergence,1024,0,527871,7.0818,0.165758056616664,408484.7453125
23,ocp,slowConvergence,2048,0,2104319,28.6943,0.698442242899631,1632267.0203125
24,ocp,order,8,0,28,0.0069,0.000875595035770913,440.303125
25,ocp,order,16,0,120,0.0076,0.000699205898780101,449.984375
26,ocp,order,32,0,496,0.0072,0.000632455532033676,464.5359375
27,ocp,order,64,0,2016,0.0084,0.00126491106406735,499.0609375
28,ocp,order,128,0,8128,0.0099,0.00128668393770792,567.1046875
29,ocp,order,256,0,32640,0.0165,0.00190029237516523,727.7859375
30,ocp,order,512,0,130816,0.0342,0.00229975844142138,1017.265625
31,ocp,order,1024,0,523776,0.0957,0.00527151675411251,2035.3125
32,ocp,order,2048,0,2096128,0.348,0.0073181661333667,4005.5234375
33,ocp,order,4096,0,8386560,1.3162,0.152907234041501,9903.06875
34,ocp,order,8192,0,33550336,3.6949,0.324936728883913,35478.8234375
35,ocp,order2,8,8184,130956,0.0624,0.00171269767715535,20231.76875
36,ocp,order2,16,8176,261720,0.0742,0.00122927259430572,20235.2640625
37,ocp,order2,32,8160,522672,0.1012,0.0019888578520235,20242.6296875
38,ocp,order2,64,8128,1042272,0.1567,0.00283039062871384,20262.8734375
39,ocp,order2,128,8064,2072256,0.259,0.00326598632371091,20294.6359375
40,ocp,order2,256,7936,4095360,0.4694,0.00699523647443981,20380.684375
41,ocp,order2,512,7680,7994112,0.8847,0.00958065179874987,20504.7140625
42,ocp,order2,1024,7168,15201792,1.6376,0.0239499478078763,21136.7796875
43,ocp,order2,2048,6144,27257856,2.8916,0.022969545054267,21903.753125
44,ocp,order2,4096,4096,41932800,4.4515,0.140523821626244,23997.1890625
45,comet,magic-simple,8,19,1401,0.121,0.00316227766016838,32768
46,comet,magic-simple,16,51,15501,0.126,0.00516397779494323,32768
47,comet,magic-simple,32,115,131904,0.162,0.00421637021355784,32768
48,comet,magic-simple,64,243,1063720,0.651,0.0185292561462497,65536
49,comet,magic-simple,128,499,8490232,13.848,0.283619463365969,65536
50,comet,magic-simple,256,1011,67729560,323.097,0.807452096962344,131072
51,comet,magic-redundant,8,3,129,0.123,0.00483045891539648,32768
52,comet,magic-redundant,16,3,361,0.122,0.00421637021355784,32768
53,comet,magic-redundant,32,3,1209,0.127,0.00483045891539648,32768
54,comet,magic-redundant,64,3,4441,0.146,0.00516397779494321,65536
55,comet,magic-redundant,128,3,17049,0.259,0.0119721899973786,65536
56,comet,magic-redundant,256,3,66841,0.922,0.0113529242439509,131072
57,comet,magic-redundant,512,3,264729,4.189,0.0255821118057999,524288
58,comet,magic-redundant,1024,3,1053721,23.017,0.0352924291535099,2097152
59,comet,slowConvergence,8,0,103,0.121,0.00316227766016838,32768
60,comet,slowConvergence,16,0,335,0.127,0.0125166555703457,32768
61,comet,slowConvergence,32,0,1183,0.128,0.00421637021355784,32768
62,comet,slowConvergence,64,0,4415,0.144,0.006992058987801,32768
63,comet,slowConvergence,128,0,17023,0.205,0.00527046276694729,65536
64,comet,slowConvergence,256,0,66815,0.493,0.0163639169448448,65536
65,comet,slowConvergence,512,0,264703,1.941,0.0260128173535023,262144
66,comet,slowConvergence,1024,0,1053695,8.474,0.0330655913803659,524288
67,comet,slowConvergence,2048,0,4204543,39.059,0.350917844136393,2097152
68,comet,order,8,0,28,0.122,0.00421637021355784,32768
69,comet,order,16,0,120,0.124,0.0126491106406735,32768
70,comet,order,32,0,496,0.122,0.00421637021355784,32768
71,comet,order,64,0,2016,0.121,0.00316227766016838,32768
72,comet,order,128,0,8128,0.122,0.00421637021355784,32768
73,comet,order,256,0,32640,0.129,0.00316227766016838,32768
74,comet,order,512,0,130816,0.14,0,32768
75,comet,order,1024,0,523776,0.19,0.0176383420737639,32768
76,comet,order,2048,0,2096128,0.322,0.00421637021355784,65536
77,comet,order,4096,0,8386560,0.916,0.0201108041719978,65536
78,comet,order,8192,0,33550336,5.474,0.0340587727318529,65536
79,comet,order2,8,8184,130956,0.249,0.0296085573216033,65536
80,comet,order2,16,8176,261720,0.25,0.00816496580927727,65536
81,comet,order2,32,8160,522672,0.283,0.0211081869319834,65536
82,comet,order2,64,8128,1042272,0.342,0.0244040069569642,65536
83,comet,order2,128,8064,2072256,0.453,0.0194650684275419,65536
84,comet,order2,256,7936,4095360,0.67,0.0205480466765632,65536
85,comet,order2,512,7680,7994112,1.094,0.0422163738639668,65536
86,comet,order2,1024,7168,15201792,1.866,0.0386436713231718,65536
87,comet,order2,2048,6144,27257856,3.225,0.0683536555146996,65536
88,comet,order2,4096,4096,41932800,5.922,0.185998805253033,65536
}
\end{center}
}
\vspace{-2mm}
\caption{Micro-benchmarks (\comet).}
\label{tab:micro2}
\vspace{-6mm}
\end{table}

Detailed numerical results (including the average peak memory
consumption) are shown in Tables~\ref{tab:micro1} and
\ref{tab:micro2}.  Note how the numbers of failures (search effort)
are identical in both solvers, and the number of propagation events are
identical for {\tt order} and usually smaller for the other
benchmarks. The main reason is the reliance on non-injective views for
the reified equalities. The running times are given in seconds and the
peak memory usage in kilobytes. Finally, it is worth highligthing that
\comet{} uses a dedicated memory allocator. The allocator relies on
block allocation through the \texttt{mmap} APIs, allocates at least 32
megabytes, and uses a grouping strategy based on block sizes. It
follows, for instance, that all domains are contiguous in memory which
leads to better cache behavior. The current implementation of \ocp{},
on the other hand, relies on \texttt{malloc} directly, which does not
exploit the locality just mentioned.  Small standard deviations on
running times are to be expected since all the benchmarks are
deterministic.

\subsection{Profiling Benchmarks}

Table~\ref{tab:profile} reports the profiling results for the
\texttt{magic-simple} benchmark of size 128 with {\sc
  Instruments}\footnote{{\sc Instruments} is Apple's version of the
  Sun MicroSytem tool {\tt DTrace}.}. {\sc Instruments} uses a
sampling-based approach to profiling. The run lasted almost 16 seconds
with a sample captured every millisecond. Each sample is a snapshot of
the runtime stack giving insights into which functions are running and
on whose behalf. The report highlights that the function for
retrieving the bounds of a variable is responsible for almost half the
runtime. The second most expensive call is method {\tt propagate} of
the linear equation propagator which accounts for 34.3\%. The third
highest is {\tt objc\_gSend}, the \textsc{Objective-C} runtime
function responsible for implementing dynamic dispatching at
4.5\%. The scheduling of closure events follows closely at 3.7\%. The
next function (an \textsc{Objective-C} closure) is the propagation
loop at 1.1\%. The remaining lines show increasingly small
contributors that include views, updates to the bounds, event
notifications to the literals (for the reified views), and the
trail-based backtracking logic.

The most important message is that dynamic dispatching is a mere 4.5\%
of the runtime on a benchmark that depends on the propagation engine
to relay events and implement views. As expected, the search is
virtually invisible from the profile as only 1,011 failures occur.

\begin{table}[tbh]
{\scriptsize
\begin{center}
\pgfplotstabletypeset[
columns = {perc,self,symb},
col sep = ampersand,
columns/rt/.style={column type=r,precision=1,fixed,column name=Time (ms)},
columns/perc/.style={column type=r,
  column name=Percentage,
      dec sep align,
        postproc cell content/.append code={
            \ifnum1=\pgfplotstablepartno
                \pgfkeysalso{@cell content/.add={}{\%}}%
            \fi
        },
        fixed,
        fixed zerofill
},
columns/self/.style={column type=r,precision=1,fixed,column name=Self (ms)},
columns/symb/.style={string type,
                     column type=l,
                     postproc cell content/.code={\edef\temp{\noexpand\lstinline\noexpand{##1}}%
                     \pgfkeyslet{/pgfplots/table/@cell content}{\temp}%
                            },
                     column name=Symbol
},
 every head row/.style={before row=\toprule%
 ,after row=\midrule},
 every last row/.style={after row=\bottomrule}
]{
rt  &perc &	self &		symb
7613.0 &   49.4 &	7613.0 &	 	bounds
5289.0 &   34.3 &	5289.0 &	 	-[CPEquationBC propagate]
693.0 &    4.5 &	693.0 &	 	objc_gSend
574.0 &    3.7 &	574.0 &	 	scheduleClosureEvt
178.0 &    1.1 &	178.0 &	 	__propagateFDM_block_invoke
116.0 &    0.7 &	116.0 &	 	-[CPLiterals changeMaxEvt:sender:]
106.0 &    0.6 &	106.0 &	 	assignTRInt
92.0 &    0.5 &	92.0 &	 	-[CPIntVarI changeMaxEvt:sender:]
88.0 &    0.5 &	88.0 &	 	-[CPIntVarI bindEvt:]
80.0 &    0.5 &	80.0 &	 	-[CPBitDom updateMax:for:]
63.0 &    0.4 &	63.0 &	 	-[CPIntVarI domEvt:]
55.0 &    0.3 &	55.0 &	 	-[CPIntFlipView bounds]
39.0 &    0.2 &	39.0 &	 	-[CPIntVarI updateMin:]
38.0 &    0.2 &	38.0 &	 	-[CPMultiCast changeMaxEvt:sender:]
30.0 &    0.1 &	30.0 &	 	-[ORTrailI backtrack:]
24.0 &    0.1 &	24.0 &	 	-[CPIntFlipView updateMin:andMax:]
21.0 &    0.1 &	21.0 &	 	objc_gSendSuper2
18.0 &    0.1 &	18.0 &	 	-[CPIntVarI bounds]
18.0 &    0.1 &	18.0 &	 	-[CPIntVarI updateMax:]
17.0 &    0.1 &	17.0 &	 	-[CPEngineI trail]
13.0 &    0.0 &	13.0 &	 	-[CPBitDom updateMin:for:]
13.0 &    0.0 &	13.0 &	 	-[CPIntVarI max]
11.0 &    0.0 &	11.0 &	 	assignTRLong
11.0 &    0.0 &	11.0 &	 	-[CPEQLitView bindEvt:]
10.0 &    0.0 &	10.0 &	 	-[CPMultiCast tracksLoseEvt:]
10.0 &    0.0 &	10.0 &	 	-[CPBoundsDom max]
}
\end{center}
}
\vspace{-2mm}
\caption{Profiling of magic-simple(128) for \ocp.}
\label{tab:profile}
\vspace{-6mm}
\end{table}

To assess the impact of views, reified equality constraints of the
form \[
b_k \Leftrightarrow x = k
\]
were substituted to refied views and the model flattening used the
traditional encoding with explicitly reified
constraints. Table~\ref{tab:profile-reified} shows the performance
profile for the same instance of {\tt magic-simple}. Naturally, the
number of choices remains the same, but the number of propagation
events increases from $4,650,800$ to $19,692,586$ for a profiling time
of 18 seconds (rather than 16 seconds). The profile shows three new
closures created inside the {\tt post} method of the {\tt
  CPReifyEqualDC} constraint: Together they implement the AC-5
protocol for each reification and account for $1.8\% + 0.9\% + 0.8\% =
3.7\%$ of the total execution time while {\tt CPDenseTriggerMap} is
the object devoted to triggers installed on values of variables as in
{\tt cc(fd)}.
Clearly, using views for reifications does improve the running time,
but even with a flood of events, the microkernel performance remains
solid.

\begin{table}[tbh]
{\scriptsize
\begin{center}
\pgfplotstabletypeset[
columns = {perc,self,symb},
col sep = ampersand,
columns/rt/.style={column type=r,precision=1,fixed,column name=Time (ms)},
columns/perc/.style={column type=r,
  column name=Percentage,
      dec sep align,
        postproc cell content/.append code={
            \ifnum1=\pgfplotstablepartno
                \pgfkeysalso{@cell content/.add={}{\%}}%
            \fi
        },
        fixed,
        fixed zerofill
},
columns/self/.style={column type=r,precision=1,fixed,column name=Self (ms)},
columns/symb/.style={string type,
                     column type=l,
                     postproc cell content/.code={\edef\temp{\noexpand\lstinline\noexpand{##1}}%
                     \pgfkeyslet{/pgfplots/table/@cell content}{\temp}%
                            },
                     column name=Symbol
},
 every head row/.style={before row=\toprule%
 ,after row=\midrule},
 every last row/.style={after row=\bottomrule}
]{
rt  &perc &	self &		symb
5953.0 &   34.5 &	5953.0 &	 	bounds
4962.0 &   28.8 &	4962.0 &	 	-[CPEquationBC propagate]
1585.0 &    9.2 &	1585.0 &	 	objc_msgSend
990.0 &    5.7 &	990.0 &	 	-[CPDenseTriggerMap loseValEvt:solver:]
576.0 &    3.3 &	576.0 &	 	-[CPTriggerMap bindEvt:]
460.0 &    2.6 &	460.0 &	 	scheduleClosureEvt
317.0 &    1.8 &	317.0 &	 	__23-[CPReifyEqualcDC post]_block_invoke54
316.0 &    1.8 &	316.0 &	 	__propagateFDM_block_invoke
193.0 &    1.1 &	193.0 &	 	-[CPBoundsDom bind:for:]
186.0 &    1.0 &	186.0 &	 	-[CPBitDom updateMax:for:]
157.0 &    0.9 &	157.0 &	 	__23-[CPReifyEqualcDC post]_block_invoke58
154.0 &    0.8 &	154.0 &	 	__23-[CPReifyEqualcDC post]_block_invoke
149.0 &    0.8 &	149.0 &	 	-[ORTrailI backtrack:]
129.0 &    0.7 &	129.0 &	 	-[CPMultiCast loseValEvt:sender:]
100.0 &    0.5 &	100.0 &	 	-[CPEngineI scheduleTrigger:onBehalf:]
86.0 &    0.4 &	86.0 &	 	-[CPIntVarI bounds]
73.0 &    0.4 &	73.0 &	 	-[CPIntVarI tracksLoseEvt:]
73.0 &    0.4 &	73.0 &	 	-[CPIntVarI changeMaxEvt:sender:]
70.0 &    0.4 &	70.0 &	 	-[CPIntVarI loseValEvt:sender:]
46.0 &    0.2 &	46.0 &	 	-[CPIntVarI bindEvt:]
44.0 &    0.2 &	44.0 &	 	-[CPIntVarI min]
42.0 &    0.2 &	42.0 &	 	objc_msgSendSuper2
41.0 &    0.2 &	41.0 &	 	assignTRInt
32.0 &    0.1 &	32.0 &	 	-[CPIntVarI bind:]
21.0 &    0.1 &	21.0 &	 	-[CPIntFlipView updateMin:andMax:]
20.0 &    0.1 &	20.0 &	 	-[CPIntFlipView bounds]
19.0 &    0.1 &	19.0 &	 	assignTRLong
19.0 &    0.1 &	19.0 &	 	-[CPBitDom remove:for:]
18.0 &    0.1 &	18.0 &	 	bool objc::DenseMapBase
18.0 &    0.1 &	18.0 &	 	-[CPMultiCast changeMaxEvt:sender:]
17.0 &    0.0 &	17.0 &	 	-[CPBitDom updateMin:for:]
}
\end{center}
}
\vspace{-2mm}
\caption{Profiling of magic-simple(128) for \ocp (Views turned off).}
\label{tab:profile-reified}
\vspace{-6mm}
\end{table}

\subsection{Application Benchmarks}

This section considers some application benchmarks to conclude the experimental study.

\paragraph{Benchmarks}

The benchmarks were selected to exercise the engine over a reasonably
broad set of constraint types. This paragraph briefly reviews each
benchmark and highlights the modeling choices

\begin{description}
\item[Golomb] This is an optimization benchmark with a blend of
  arithmetic and global constraints (i.e.,
  \texttt{alldifferent}). Both solvers enforce domain consistency on
  the {\tt alldifferent} constraints. The arithmetic constraints
  contain a mix of equalities and inequalities of low arity. The size
  was picked to have a very long running test.

\item[Knapsack] An optimization benchmark where each knapsack relies
  on the global constraint. The objective function is simply to
  maxinimize the profit.

\item[Perfect] This is a constraint satisfaction benchmark. It uses
  arithmetic constraints and logical constraints to state the
  non-overlapping requirements and to demand that the squares cutting
  through any horizontal (vertical) cut line fit exactly within the
  container. Its labeling first focuses on the $x$ axis attempting to
  choose, for each abscissa, a square to pin at that location. It
  carries on with an identical process over the $y$ axis.

\item[PPP] This is a constraint satisfaction benchmark. It uses
  \texttt{alldifferent}, \texttt{packing}, reified and arithmetic
  constraints to impose all the requirements. The labeling focuses on
  the earliest periods first and uses a static variable ordering
  within each period when considering each guest.

\item[Steel Mill] The is an optimization problem where the objective
  function uses element constraints to aggreggate the losses incurred
  on each slab and the requirements are expressed with
  \texttt{packing}, arithmetic and logical constraints. The search
  scans the slab using the first fail principle and uses a dynamic
  symmetry breaking for the value labeling to avoid considering more
  than one unused value.

\item[Sport] This a constraint satisfaction problem that uses several
  global constraint types. It relies on {\tt alldifferent}, {\tt
    cardinality}, and {\tt table} constraints (all using domain
  consistency) alongside with static symmetry breaking (ordering the
  home-away variables for each period and game).
\end{description}

Each benchmark used a search heuristic whose behavior is identical for
both \ocp{} and \comet{} and was evaluated over a series of 10 runs on
both systems. Three sets of benchmarks (\texttt{knapsack, golomb
  ruler, perfect square}) produce exactly the same dynamic search tree
whereas the remaining benchmarks exhibit some differences in the
number of choices due to randomization.  The objective in selecting
the search heuristic was not to get the absolute best-known result for
each benchmark but to ensure that both systems were performing the
same search (possibly modulo some randomization) in order to focus on
the the propagation engine.

Table~\ref{tab:ocp} reports the results for \ocp{} (left) and \comet{}
(right) on 7 benchmarks: The Golomb ruler (size 13), the optimization
version of the knapsack problem (instance 3), the progressive party
benchmark (parameters 1,9), the Steel Mill Slab Design problem with
symmetry breaking, the perfect square problem, and sport
scheduling. For each benchmark, the table reports the average CPU time
and wall-clock time, the standard deviation over computing times, the
number of choices, and the peak memory consumed (in megabytes).

\begin{table}[t]
{\scriptsize
\begin{center}
\noindent\pgfplotstabletypeset[
  columns={name,size,cpu,sigcpu,wc,nchoice,mpeak},
  string replace*={psq}{perfect},
  columns/name/.style ={string type,column name=$Bench$,column type=l|},
  columns/size/.style ={string type,column name=$sz$,column type=l|},
  columns/cpu/.style={column name=$\mu(T_{cpu})$,column type=r,precision=2,fixed},
  columns/sigcpu/.style={column name=$\sigma(T_{cpu})$,column type={p{1cm}},precision=2,fixed},
  columns/wc/.style={column name=$\mu(T_{wc})$,column type=r|,precision=2,fixed},
  columns/nchoice/.style={column name=$Choices$,column type=r,fixed},
  columns/nfail/.style={column name=$Failures$,column type=r|,precision=1,fixed},
  columns/nprop/.style={column name=$Prop$,column type=r|,precision=1,fixed},
  columns/mused/.style={column name=$|M|$,column type=r,precision=2,preproc/expr=##1 / 1024,fixed},
  columns/mpeak/.style={column name=$|P|$,column type=r,precision=2,preproc/expr=##1 / 1024,fixed},
  col sep=comma,
  every head row/.style={before row=\toprule%
  ,after row=\midrule},
  every last row/.style={after row=\bottomrule}
]{
,solver,name,size,cpu,wc,nchoice,nfail,nprop,mused,mpeak,tpc,sigwc,sigcpu
1,ocp,golomb,8,0.028,0.0316,595,582,41440,379.6515625,557.2453125,4.70588235294118e-05,0.00236643191323985,0.00210818510677892
2,ocp,golomb,9,0.1533,0.1575,3149,3131,295777,382.825,598.16875,4.86821213083519e-05,0.00287711275027201,0.00312872000806862
3,ocp,golomb,10,1.1355,1.1406,19938,19920,2317044,386.2421875,623.1484375,5.69515498043936e-05,0.0139459273226591,0.0138021737418422
4,ocp,golomb,11,24.4527,24.4712,321438,321411,50288996,376.6203125,708.9171875,7.60728351968342e-05,0.474943458623118,0.475077315345151
5,ocp,golomb,12,243.7386,243.8703,2656760,2656729,487732100,395.934375,753.45,9.1742799500143e-05,0.861565635598617,0.864697403719936
6,ocp,golomb,13,4238.6962,4241.659,35497998,35497967,-333202615,597.1015625,1027.7890625,0.000119406626818786,32.9707876527761,31.4094376680357
7,ocp,ks,1,0.0762,0.0903,105,70,62280,5906.7671875,7671.3890625,0.000725714285714286,0.00388873015549064,0.00379473319220206
8,ocp,ks,2,0.5294,0.5501,890,832,739563,10032.8734375,13036.0453125,0.000594831460674157,0.01842974648647,0.0178649252882724
9,ocp,ks,3,19.9606,20.034,20867,20745,16877120,26321.88125,33892.840625,0.000956562994201371,0.175188787058736,0.174163524692552
10,ocp,ks,4,80.2256,80.4024,307268,306789,66670552,23725.3765625,31070.6578125,0.000261093247588424,0.540637113865566,0.510822039026855
11,ocp,sport,0,4.6507,4.6567,19637,19584,3328400,1465.7578125,1900.4765625,0.000236833528543057,0.0163166581545774,0.0163302718219195
12,ocp,ppp,0,0.6244,0.6398,5181,5000,1513442,6750.9796875,15053.8546875,0.000120517274657402,0.0131807941086011,0.0124293559321829
13,ocp,slab,0,3.4392,3.4742,3005,1841,1580131,11148.9546875,36725.7984375,0.0011444925124792,0.0134064661015994,0.0133649624848789
14,ocp,psq,0,6.5351,6.5692,125809,125802,27676628,4830.921875,25777.921875,5.19446144552457e-05,0.0812742955128564,0.0817114843009638
}
\noindent\pgfplotstabletypeset[
  columns={name,size,cpu,sigcpu,wc,nchoice,mpeak},
  string replace*={psq}{perfect},
  columns/name/.style ={string type,column name=$Bench$,column type=l|},
  columns/cpu/.style={column name=$\mu(T_{cpu})$,column type=|r,precision=2,fixed},
  columns/sigcpu/.style={column name=$\sigma(T_{cpu})$,column type=r,precision=2,fixed},
  columns/wc/.style={column name=$\mu(T_{wc})$,column type=r|,precision=2,fixed},
  columns/nchoice/.style={column name=$Choices$,column type=r,fixed},
  columns/nfail/.style={column name=$Failures$,column type=r|,precision=1,fixed},
  columns/nprop/.style={column name=$Prop$,column type=r|,precision=1,fixed},
  columns/mused/.style={column name=$|M|$,column type=r,precision=2,preproc/expr=##1 / 1024,fixed},
  columns/mpeak/.style={column name=$|P|$,column type=r,precision=2,preproc/expr=##1 / 1024,fixed},
  col sep=comma,
  every head row/.style={before row=\toprule%
  ,after row=\midrule},
  every last row/.style={after row=\bottomrule}
]{
,solver,name,size,cpu,wc,nchoice,nfail,nprop,mused,mpeak,tpc,sigwc,sigcpu
1,comet,golomb,8,0.163,0.278,595,588,43745,25194,32768,0.000273949579831933,0.250368617131709,0.00948683298050514
2,comet,golomb,9,0.282,0.318,3149,3140,309862,25484,32768,8.95522388059702e-05,0.00632455532033676,0.00632455532033674
3,comet,golomb,10,1.227,1.264,19938,19929,2436692,25467,32768,6.15407764068613e-05,0.0250333111406915,0.0240601099101581
4,comet,golomb,11,23.673,23.739,321438,321425,53427294,32503,65536,7.36471730162582e-05,0.136255478992793,0.134168219452713
5,comet,golomb,12,236.297,236.611,2656760,2656744,524021274,31600,65536,8.89417937638326e-05,0.841433300981128,0.829766366046625
6,comet,golomb,13,4065.351,4071.155,35497998,35497987,340217140,28637,65536,0.000114523388051349,29.6430922251149,28.9984805923805
7,comet,ks,1,0.214,0.257,105,74,73629,32295,65536,0.00203809523809524,0.0105934990547138,0.00843274042711568
8,comet,ks,2,0.738,0.785,890,834,794007,36525,65536,0.000829213483146067,0.0126929551764399,0.0122927259430572
9,comet,ks,3,28.05,28.204,20867,20750,18003308,59000,131072,0.00134422772799157,0.190857713144286,0.189443629845104
10,comet,ks,4,96.522,96.864,307268,306822,76952979,57486,131072,0.000314129684835388,0.992037185682965,0.970461516782377
11,comet,sport,0,5.001,5.046,23523,23465,4124178,25893,32768,0.000212600433618161,0.0333999334663343,0.0347850542618521
12,comet,ppp,0,0.633,0.678,5607,5425,2001894.5,36552,65536,0.000112894596040663,0.0175119007154182,0.0156702123647242
13,comet,slab,0,5.689,5.797,3206,2248,15107084.4,77014,131072,0.00177448533998752,0.111060544048931,0.104716548623203
14,comet,psq,0,6.634,6.726,125809,251618,14131577,58863,131072,2.63653633682805e-05,0.114328960071852,0.11374435272917
}
\end{center}
}
\vspace{-2mm}
\caption{Performance test for \ocp{} (Top) and \comet{} (Bottom).}
\label{tab:ocp}
\vspace{-6mm}
\end{table}

\begin{figure}[t]
\includegraphics[width=\textwidth]{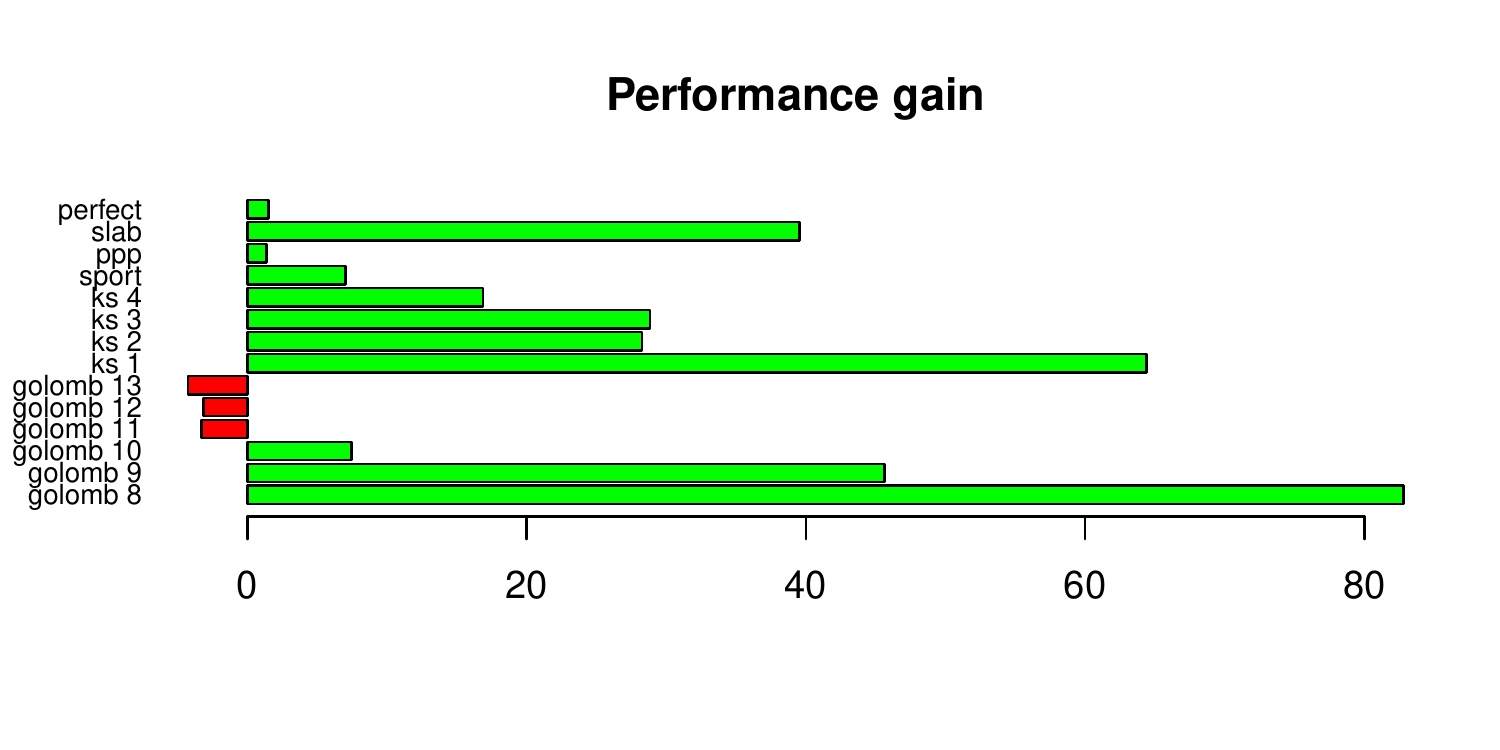}
\vspace{-15mm}
\caption{Performance Gains (Losses) of Objective-CP versus
  \textsc{Comet}.}
\label{fig:ratio}
\end{figure}

\paragraph{Running Time}

Figure~\ref{fig:ratio} offers a quick overview of the performance
ratio between the two implementations. Each bar is a percentage
established as
\[
\frac{\mu(T_{cpu}^{\mbox{\textsc{Comet}}}) - 
\mu(T_{cpu}^{\mbox{\textsc{Objective-CP}}})}
{\mu(T_{cpu}^{\mbox{\textsc{Comet}}})}\cdot 100.0
\]
It shows that the \textsc{Objective-CP} implementation is generally
competitive with \textsc{Comet} when comparing running times.  A
detailed view appears in Table~\ref{tab:ocp}. The main conclusion
drawn from the results is that, despite its generality, the
preliminary status of the implementation, and the reliance on dynamic
dispatching within the implementation, the microkernel of \ocp{} is
competitive with the polished \comet{} implementation.
The \textit{narrow} loss on the larger Golomb instances appear to be
due to the difference in memory management style. \comet{}, with its
dedicated allocator, clusters all the objects of identical sizes in
contiguous regions of virtual memory leading to good cache
behavior. \ocp{} currently relies exclusively on \texttt{malloc} and
seems to suffer slightly from that choice. The observation was
confirmed with \texttt{DTrace} that shows a larger volume of L3 cache
misses per time unit.

\paragraph{Memory Consumption}

The memory behavior of \textsc{Objective-CP} shows a significant
improvement over \textsc{Comet}. This is easily explained as the
former is based on a thin object-oriented layer on top of {\tt C}
whereas \comet{} relies on a compiler and a just-in-time code
generation that both add some overhead. Nonetheless, the gains are so
significant that they are worth highlighting. Column $|P|$ gives the
peak memory consumptions (in megabytes) as reported by \texttt{malloc}
for \ocp{} and by the garbage collector library for \comet{}. The peak
usage memory footprint drops by up to a factor of 60 on {\tt golomb},
at least a factor of 4 on the largest knapsack instance. Overall,
\ocp{} exhibits frugal memory needs. Finally, it is worth remembering
that the entire implementation adopts the reference counting strategy
of the underlying \textsc{NextStep Foundation} libraries rather than a
garbage collector.

\section{Conclusion}
\label{sec:ccl}

This paper presented a microkernel architecture for a new constraint
programming solver. The microkernel strives to offer a minimal API
which remains domain agnostic and facilitates the construction of any
domain-specific engine as a service on top of the microkernel. The
paper showed that such a microkernel can be built for constraint
programming and provides a small but versatile set of functionalities.
Moreover, the resulting microkernel can be implemented to be
competitive with state-of-the-art monolithic solvers.

\section*{Acknowledgments}

We would like to express our gratitude to Thibaux Freydy and Peter
Stuckey for many interesting discussions. In particular, Thibaux
encouraged us to eliminate the explicit handling of failures in
propagators. NICTA is funded by the Australian Government as
represented by the Department of Broadband, Communications and the
Digital Economy and the Australian Research Council through the ICT
Centre of Excellence program.

\bibliographystyle{plain}
{\footnotesize

}

\end{document}